\documentclass{article}

    \usepackage[preprint]{neurips_2025}

\usepackage[utf8]{inputenc} % allow utf-8 input
\usepackage[T1]{fontenc}    % use 8-bit T1 fonts
\usepackage{hyperref}       % hyperlinks
\usepackage{url}            % simple URL typesetting
\usepackage{booktabs}       % professional-quality tables
\usepackage{amsfonts}       % blackboard math symbols
\usepackage{nicefrac}       % compact symbols for 1/2, etc.
\usepackage{microtype}      % microtypography
\usepackage{xcolor}         % colors
\usepackage{array}
\usepackage{graphicx}
\usepackage{amsmath}
\usepackage{algorithm}
\usepackage{algorithmic}
\usepackage{longtable}
\usepackage[normalem]{ulem}
\newcolumntype{H}{>{\setbox0=\hbox\bgroup}c<{\egroup}@{}}
\hypersetup{
     colorlinks = true,
     linkcolor = red,
     anchorcolor = blue,
     citecolor = teal,
     filecolor = blue,
     urlcolor = black
     }

\title{\textsc{GP-MoLFormer-Sim}: Test Time Molecular Optimization through Contextual Similarity  Guidance}

\author{%
  Ji\v{r}\'\i\,Navr\'atil\thanks{Equal contribution} \\
  IBM Research \\
  \texttt{jiri@us.ibm.com}
  \And
  Jarret Ross\footnotemark[1] \\
  IBM Research \\
  \And
  Payel Das\\
  IBM Research \\
  \texttt{daspa@us.ibm.com}
  \And
    Youssef Mroueh\\
  IBM Research \\
  \And
    Samuel C Hoffman\\
  IBM Research \\
  \And
    Vijil Chenthamarakshan\\
  IBM Research \\
  \And
    Brian Belgodere\\
  IBM Research \\
}

\begin{document}

\maketitle

\begin{abstract}
    The ability to design molecules while preserving similarity to a target molecule and/or property is crucial for various applications in drug discovery, chemical design, and biology. We introduce in this paper an efficient training-free method for navigating and sampling from the molecular space with a generative Chemical Language Model (CLM), while using the molecular similarity to the target as a guide. Our method leverages the contextual representations learned from the CLM itself to estimate the molecular similarity, which is then used to adjust the autoregressive sampling strategy of the CLM. At each step of the decoding process, the method tracks the distance of the current generations from the target and updates the logits to encourage the preservation of similarity in generations. We implement the method using a recently proposed $\sim$47M parameter SMILES-based CLM, \textsc{GP-MoLFormer}, and therefore refer to the method as \textsc{GP-MoLFormer-Sim}, which enables a test-time update of the deep generative policy to reflect the contextual similarity to a set of guide molecules. The method is further integrated into a genetic algorithm (GA) and  tested on a set of standard molecular optimization benchmarks involving property optimization, molecular rediscovery, and structure-based drug design. Results show that, \textsc{GP-MoLFormer-Sim}, combined with GA (\textsc{GP-MoLFormer-Sim+GA}) outperforms  existing  training-free  baseline methods, when the oracle remains black-box. The findings in this work are a step forward in understanding and guiding the generative mechanisms of CLMs. 
\end{abstract}

\section{Introduction}
Finding new functional molecules with desired structure and properties involves solving  a constrained multi-objective optimization problem, which is crucial in many applications such as drug discovery and new material design. Given the large size of the molecular space, brute-force search around known substructures is often inefficient and costly for such tasks. 
Existing molecular optimization algorithms therefore mainly involve reinforcement learning, deep generative models, genetic algorithms, or a combination thereof. Recent works show that traditional genetic algorithm (GA)-based methods with domain-specific operators are competitive when compared to costlier alternatives that involve deep learning models \cite{gao2022sample, tripp2023geneticalgorithmsstrongbaselines}. Earlier efforts that have successfully combined GAs with deep learning for better search typically require further training of the deep learning model, more specifically of the deep generative model, to adapt the generative policy for generating high-reward samples corresponding to the specific optimization problem \cite{ahn2020guiding, kim2024geneticguidedgflownetssampleefficient}.  

Different from earlier approaches, here we propose a \textit{training-free} method for equipping a pre-trained deep generative model for targeted search. The proposed method exploits the contextual similarity between a target molecule and a set of generated molecules with a generative chemical language model (CLM).  We update the autoregressive decoding policy of the generative model on the fly as a means to guide the generation toward high-reward samples. We use the recently proposed SMILES-based \textsc{GP-MoLFormer} model \cite{ross2025gpmolformerfoundationmodelmolecular} as the base generative CLM to generate molecules, and therefore refer to this test-time contextual similarity-based guided generation method as \textsc{GP-MoLFormer-Sim}. We first show the performance of \textsc{GP-MoLFormer-Sim} on a similarity-based lead optimization task where the goal is to generate molecules of high similarity with respect to a given target molecule in a sample-efficient manner. Experiments on this task show that the proposed method outperforms random search as well as a reinforcement learning-based baseline. 
 
\begin{figure}[t]
    \centering
    \includegraphics[width=\textwidth]{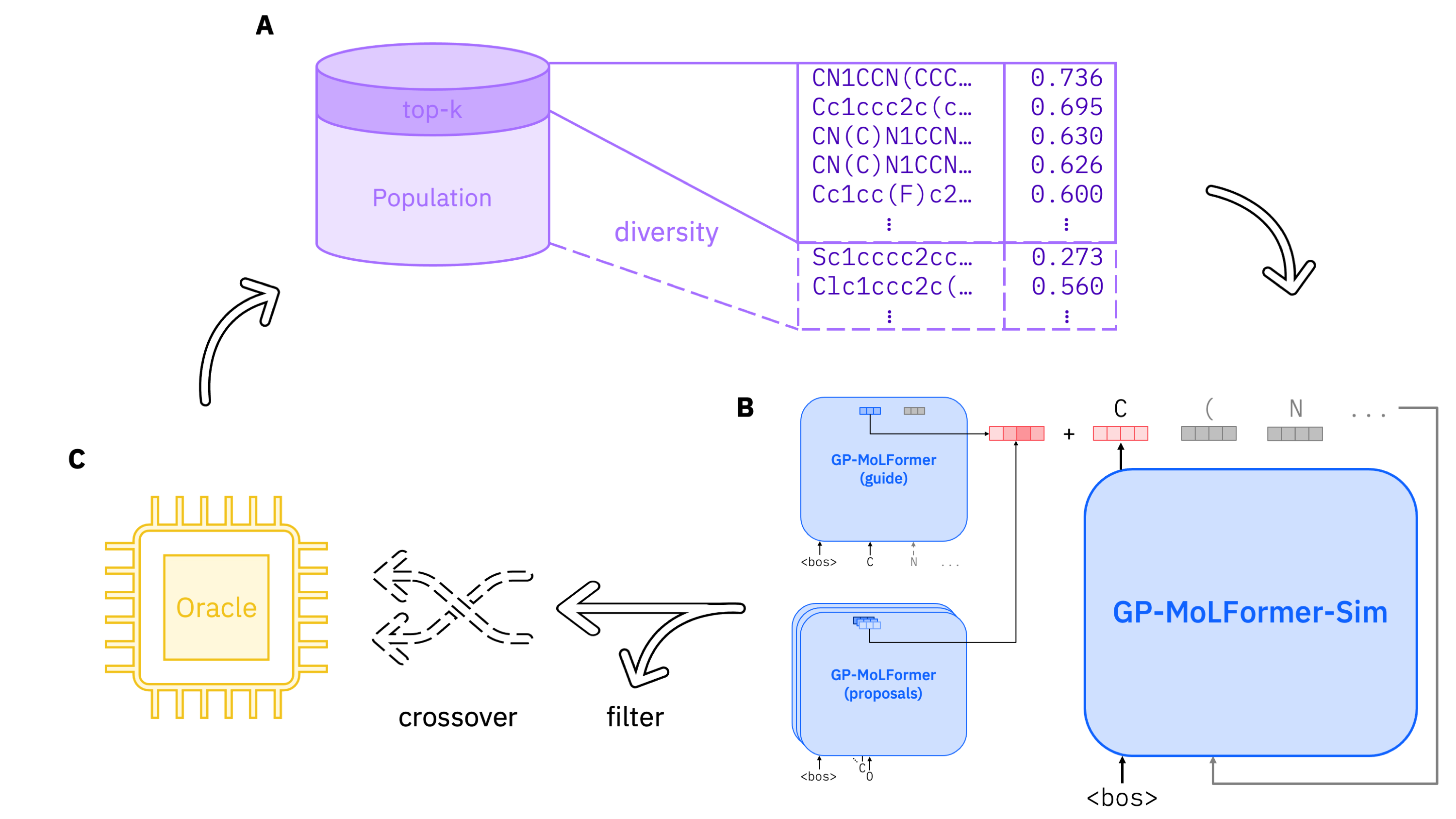}
    \caption{Overview of the \textsc{GP-MoLFormer-Sim+GA} process. (A) The top-k highest scoring molecules so far are chosen as guides with additional diverse candidates, if desired. (B) Using \textsc{GP-MoLFormer-Sim}, generate new candidates conditioned on closeness to top guides. \textsc{GP-MoLFormer-Sim} adjusts the logits of the base model at every iteration using embedding similarity of the guide sequence so far to each proposed next token. (C) Prune (filter) or augment (with graph-based crossover operation) generations and score them with the oracle. These new samples are added back to the population and the process is repeated until the oracle budget is met.}
    \label{fig:GA}
\end{figure}

We further integrate \textsc{GP-MoLFormer-Sim} with a genetic algorithm-based search process, where \textsc{GP-MoLFormer-Sim} enables generating offspring of the high-reward samples (see Figure \ref{fig:GA}). We refer to this approach as \textsc{GP-MoLFormer-Sim+GA}. Results on the popular Practical Molecular Optimization (PMO) benchmark \cite{gao2022sample} show that \textsc{GP-MoLFormer-Sim+GA} yields better performance on 23 molecular optimization tasks when the oracle is a black-box, compared to the current state-of-the-art GA-based training-free baselines including the ones that call large language models like GPT-4 for proposing high-reward samples.  To our knowledge, this is the first demonstration of using test-time update of a CLM-based deep generative policy for %incorporated into GA frameworks
for molecular optimization.  %consist of two main components - a molecular generation strategy and an optimization method.  %To circumvent the black-box nature of the evaluations that prohibits access to the gradients, 

\section{Guided Generation}

\subsection{Background information --- \textsc{GP-MoLFormer}} \textsc{GP-MoLFormer} is a chemical language foundation model, which is a GPT-style autoregressive decoder trained  with linear attention and rotary embeddings \cite{ross2025gpmolformerfoundationmodelmolecular}\footnote{Available via https://huggingface.co/ibm-research/GP-MoLFormer-Uniq}. The model used in here is trained on $\sim$650M canonicalized SMILES obtained from ZINC and PubChem databases.  Unconditional sampling from this chemical language model would allow exploring the chemical space. For details of the \textsc{GP-MoLFormer} model and its performance on unconditional SMILES generation task, see \cite{ross2025gpmolformerfoundationmodelmolecular}.

\subsection{Target-guided generation with \textsc{GP-MoLFormer} --- \textsc{GP-MoLFormer-Sim}}
\label{sec:guided_generation}
Guiding the autoregressive sampling from CLMs like \textsc{GP-MoLFormer} towards specific molecules is of a paramount interest, as it enables  generating new variations given target molecules of importance. Specifically, given a single molecule we are interested in exploring the molecular neighborhood where the similarity is defined through the cosine in the embedding space of the same generative model (\textsc{GP-MoLFormer}).

\begin{algorithm}[t]
%\label{alg:general}
\begin{algorithmic}[1]
\REQUIRE $\textsc{GetEmbeddingGPT}() $ , 
$\textsc{GetLogits}()$
\STATE \textbf{Inputs:} $\alpha$ (mixing parameter), $\tau$ (softmax temperature), $T$ (RFF)
\STATE $s\gets \text{BOS} $
\STATE $t\gets 1$
\WHILE{EOS is not met}
\STATE \textbf{Append and Embed Generated } 
\FOR{$i \in \text{Vocab}$ } 
\STATE $x_i= \textsc{GetEmbeddingGPT}(s \oplus i) $
\STATE $x_{i} \gets RandomFeatures(x_i)$ \hspace{1cm} // Optional
\STATE $x_{i} \gets \frac{x_i}{||x_i||}$
\ENDFOR
\STATE \textbf{Embed targeted molecule up to time $t$}
\FOR{ $j\in \text{targetMolecules}$ }
\STATE $ y_j= \textsc{GetEmbeddingGPT}(m_j [1:t])$
\STATE $y_{j} \gets RandomFeatures(y_j)$ \hspace{1cm} // Optional
\STATE $y_j\gets\frac{y_j}{||y_j||}$ 
\ENDFOR
\STATE \textbf{Compute pairwise cosine $(\text{Vocabsize} \times N $ where $N=\text{number of targetMolecules})$ }
\STATE $ S_{ij} = \langle x_i, y_j \rangle, i=1\dots \text{Vocabsize}$, $ j=1\dots N$
\STATE $ \bar{S}_i = \frac{1}{N}\sum_{j=1}^N S_{ij} $, for $i=1\dots \text{Vocabsize}$
\STATE \textbf{Tilting the logits}
\STATE $u \gets  \textsc{GetLogits}(s)$ (vector of size $\text{Vocabsize}$, if topk used this is k)
\STATE  Standardize $u$ and $\bar{S}$
\STATE $u \gets \frac{1}{\tau}  ( (1-\alpha) u + \alpha  \bar{S})  $
\STATE Sample with probability $\textsc{softmax}(u)$ and get token $d$ 
\STATE  $s \gets s \oplus d $
\STATE $t\gets t +1$
\ENDWHILE
\RETURN $s$
\end{algorithmic}
\caption{Guided Generation with kernel approximation }
\label{alg:guided_generation}
\end{algorithm}

More formally, we wish to generate a new sequence $s$ with guidance from molecules $m_j,j=1\dots N$ that are canonical SMILES sequences. We build the sequence $s$ incrementally by gradually sampling tokens from a new policy that mixes the likelihood under \textsc{GP-MoLFormer} (logit $u$) and the \textit{contextual} similarity of the new sequence to target molecules in the embedding space of \textsc{GP-MoLFormer}, $\bar{S}$.   Algorithm \ref{alg:guided_generation} summarizes this procedure. The logits of the new guiding policy are $\frac{1}{\tau} ( (1-\alpha)u + \alpha \bar{S})$, where $\alpha \in [0,1]$ controls the mixing  strength, interpolating between unconditional sampling from \textsc{GP-MoLFormer} and pure similarity based sampling, and $\tau$ is a sampling temperature that allows control over the entropy of the sampling. The sequence is generated iteratively until the \texttt{<eos>} token is selected.

Note that the cosine similarity is used to ``tilt'' the logits of the \textsc{GP-MoLFormer} using similarity to a neighborhood formed by target molecules in the embedding space. We can push this idea further using a kernel density estimator (KDE) with a gaussian kernel.  The temperature $T$ of the kernel  induces further locality control. We can approximate the KDE  using Random Fourier Features \citep{rff} (lines 8 and 14 in Algorithm \ref{alg:guided_generation}).

Because our proposed algorithm is a test time guidance algorithm, it does not need any training procedure and enjoys multiple advantages of scalability, parallelism, efficiency, and versatility in its applicability to multiple domains.  The complexity of the algorithm is linear in the vocabulary size  (2362), the dimensionality of \textsc{GP-MoLFormer} embedding (768) and the number of target molecules.  For experiments with random features, we also used $768$ random features.  As our method is training-free, we do not require any computational resources or timing observations for training. A single A100 GPU is used for each inference task (we also note that GPU memory is not a concern as our memory footprint only occupied no more than 8 GB of VRAM).  Even though our guided method has many more moving parts and occupies much more memory than the base model unconditional generation, the extra computation cost of the guided method is only roughly four times slower than the unconditional generation. As an example, generating a single token from the unconditional model takes, on average, $0.013$ seconds while generating a guided token takes $0.049$ seconds. When generating a batch of 20 molecules, the runtime is $1.02$ seconds (producing avg. molecule length of 43 tokens) for unconditional and $3.97$ seconds for guided generation (of molecules with 40  tokens on average).

\begin{figure}[t]
    \centering
    \includegraphics[width=1.\textwidth]{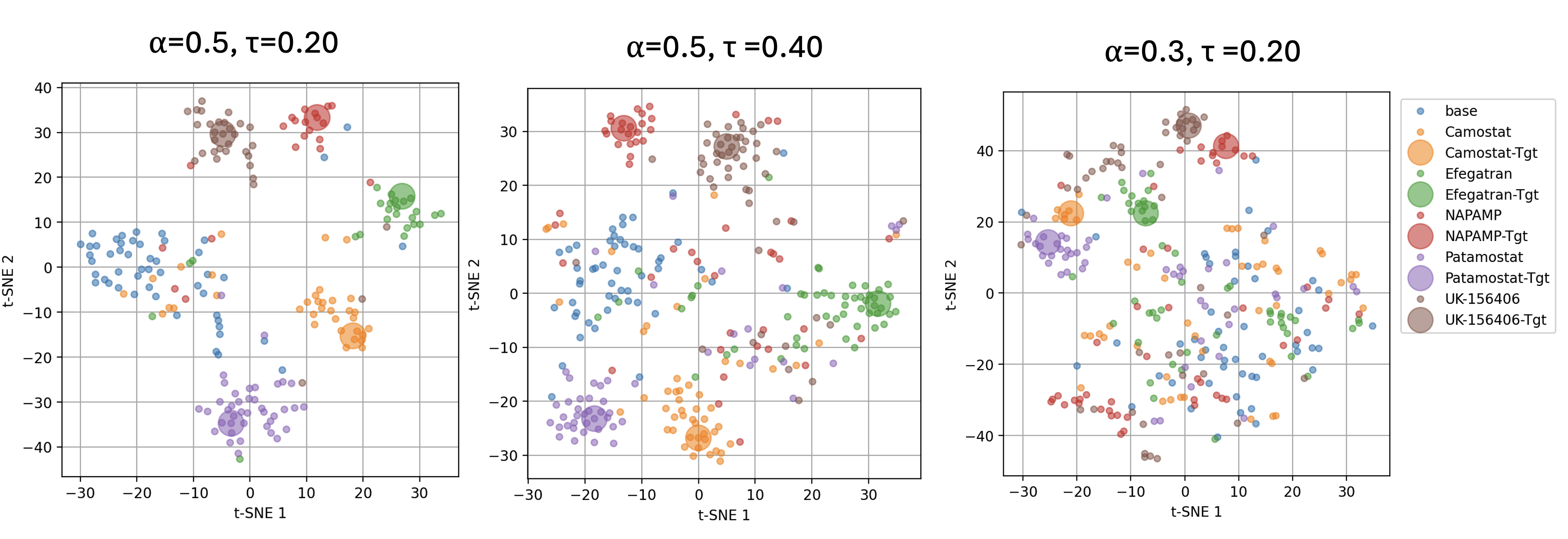}
    \caption{Guided generations (dots) around five trypsin inhibitor targets (large circles) with various guiding strength $\alpha$ and sampling temperature $\tau$ settings.  ``Base'' molecules from un-guided generation. Green, orange, red, purple and brown correspond to five different targets. Molecules are visualized in \textsc{GP-MoLFormer} embeddings space projected onto two t-SNE dimensions. } 
    \label{fig:trypsin_clusters}
\end{figure}

In Figure \ref{fig:trypsin_clusters}, we showcase a depiction of the Algorithm \ref{alg:guided_generation} in action on a guided generation task (for details of the task, see Section \ref{experiments}), which aims to generate molecules within the individual neighborhood of five trypsin inhibitor targets (large circles) with varying guiding strength $\alpha$ and sampling temperature $\tau$ settings. In blue, we see the unconditional generation from \textsc{GP-MoLFormer}. When comparing the first two panels for the same guidance strength $\alpha$, we see the effect of sampling temperature $\tau$: the higher temperature ($\tau=0.40$) leads to a higher entropy resulting in a larger spread around the target molecules. On the other hand, comparing the outer two panels, for fixed sampling temperature $\tau=0.20$, we observe that larger mixing $\alpha=0.5$ leads to tighter clustering around the target molecules, away from the unconditional baseline in blue. It should be mentioned that the algorithm is not specific to using a single guide, and can be extended to guiding the generation to multiple target molecules simultaneously. Additional visualizations involving a guidance by multiple targets simultaneously can be found in the Appendix Figure \ref{fig:trypsin_multiguide_clusters}.

\paragraph{Theoretical Justification} 
It can be shown that the computations on lines 23--24 in Algorithm \ref{alg:guided_generation}
% , where the new guided probability is obtained using a softmax on the interpolation between the KDE estimator of the target molecules and the logits of \textsc{GP-MoLFormer}, 
represent a solution of a closed-form optimization problem involving token similarity to the target molecule distribution and a KL divergence from the smoothed \textsc{GP-MoLFormer} distribution. Full detail is laid out in Appendix \ref{sec:theoretical_justification}.

\subsection{\textsc{GP-MoLFormer-Sim} augmented with genetic algorithm --- \textsc{GP-MoLFormer-Sim+GA}}
\label{sec:gpmfs-ga}
\begin{algorithm}[t]
\caption{GP-MoLFormer-Sim+GA}
\label{alg:GA}
\begin{algorithmic}[1]
\REQUIRE Oracle $F: \mathbb{M} \rightarrow \mathbb{R}$, \textsc{GP-MoLFormer-Sim}()
\STATE Inputs: $G, K, B, D$
\STATE generation $\leftarrow$ 0, budget $\leftarrow$ 0
\STATE Initialize $P \sim \mathrm{ZINC}$
\STATE Store scores $R[P] \leftarrow F(P)$
\STATE Record (budget spent, generation, avg. $K$ best scores)
\WHILE{Oracle budget $\leq$ B}
    \STATE Sample $S \subset P$ \hspace{1cm} // Select $G$ best candidates as guides
    \STATE Optional: $S \leftarrow S \cup (S' \subset P)$  \hspace{1cm} //  Augment by $D$ diverse candidates  
    \STATE Generate $N \leftarrow \textsc{GP-MoLFormer-Sim}(s) \quad\forall s \in S$ // Use GP-MoLFormer-Sim to create neighbors (mutations) of guides
    \STATE Select $P'\subset P \cup N$   \hspace{1cm} //  e.g, best by $T_{sim}$ to current top in $P$
    \STATE Optional: augment $P'$    \hspace{1cm} // e.g., via ``graph-based'' crossover of current best guides
    \STATE generation $\leftarrow$ generation + 1
    \STATE Store $R[P' \setminus P] \leftarrow F(P' \setminus P)$
    \STATE Set $P \leftarrow P'$
    \STATE Record (budget spent, generation, avg. $K$ best scores)
\ENDWHILE
\RETURN Array of tuples (generation, budget spent, avg. top-$K$ score)
\end{algorithmic}
\end{algorithm}

Typical GA combines {\em mutation} and {\em crossover} steps to augment a current candidate pool to aid exploration, followed by sampling the fittest (highest-scoring) compounds (as per the black-box oracle function) to form the next generation in a cyclical process. In our work, we adopt the cyclical nature of a GA and combine it with the ability of the \textsc{GP-MoLFormer-Sim} method to produce novel molecules with high efficiency that are close to targets already known to have a desirable property. The process is illustrated in Figure \ref{fig:GA} and captured in Algorithm \ref{alg:GA}. In every generation, we maintain a set of compounds with known property of interest --- the oracle value. The best $G$ candidates are selected to serve as guides for \textsc{GP-MoLFormer-Sim} (going from A$\rightarrow$B in Figure~\ref{fig:GA}). The selection process takes into account high oracle scores as well as diversity. For each of the $K$ guides, the \textsc{GP-MoLFormer-Sim} module (Fig.~\ref{fig:GA} B) generates a set of novel candidates forming new \emph{mutated} offspring. In order to reduce oracle budget expenditure, a pruning step (``filter'' in Figure~\ref{fig:GA} B$\rightarrow$C) is applied to reduce the offspring set size by removing candidates that are below a certain threshold of Tanimoto similarity ($T_{sim}$) measured from the current guide set (the details are given in the Appendix). Optionally,  a graph-action based crossover operation \cite{jensen2019graph} is also applied to create offspring from best guides. The offspring set is then sent to oracle for scoring and is merged to the compound pool (Fig.~\ref{fig:GA} C$\rightarrow$A), thus closing the GA cycle. During the process, the average of top-10 scoring compounds are recorded along with oracle budget expenditure. %The latter is later used to calculate the area under curve metric, as described in Section \ref{XXX}. 
A sample optimization trajectory visualized in a 2D t-SNE chart can be found in
Appendix Fig.~\ref{fig:albuterol_opt_traj} and specific GA parameter settings used are listed in Appendix Table \ref{tab:hyperparameters}.

\section{Related Work}

\textbf{Molecule optimization} The goal of molecule optimization is to iteratively modify molecule structures to improve desired properties like binding affinity, solubility, drug likeliness, etc. The ability to represent molecules using text-based encodings like SMILES \cite{weininger1988smiles} and SELFIES \cite{Krenn_2020_selfies}, enables the application of natural language processing techniques to tackle this problem. Existing methods have used techniques such as reinforcement learning (RL) \cite{olivecrona2017molecular, blaschke2018application, loeffler2024reinvent, segler2017generating, neil2018exploring, zhou2019optimization}, variational autoencoders~\cite{gomez2018automatic, jin2018junction}, Bayesian optimization~\cite{moss2020boss, tripp2021fresh}, GFlowNets~\cite{shen2023tacogfn, bengio2021flow, bengio2023gflownet}, genetic algorithms~\cite{lee2023drug}, query-based optimization~\cite{hoffman2022optimizing}, and diffusion models~\cite{lee2023exploring}. Recently, large language model-based methods~\cite{ye2025drugassist, wang2024efficient} have appeared as a promising method for molecule design, when used in combination with other methods like genetic algorithms~\cite{wang2024efficient}.

\textbf{Genetic algorithms for molecular optimization}
Genetic algorithms have emerged as a state-of-the-art method for molecule optimization tasks~\cite{gao2022sample, tripp2023geneticalgorithmsstrongbaselines}. They work by mimicking an iterative evolutionary process using operations like mutation and crossover on the molecule representations and allowing favorable candidates to survive to the next generation. Some examples include STONED~\cite{Nigam_Pollice_Krenn_Gomes_Aspuru-Guzik_2021} which operates on SELFIES representations,  GEAM~\cite{lee2023exploring} which operates on molecule fragments, Graph-GA~\cite{jensen2019graph} which applies graph-based mutation/crossover operators for GA, Mol-GA~\cite{tripp2023geneticalgorithmsstrongbaselines} which incorporates quantile uniform sampling to maintain diversity while rewarding the best candidates,  MOLLEO~\cite{wang2024efficient} which uses Graph-GA in combination with a large language model, genetic guided GFlowNets~\cite{kim2024geneticguidedgflownetssampleefficient}, and SynNet which incorporates synthesis constraints~\cite{gao2021amortized}. The present work differs from those earlier ones, as it combines a test-time guided generation using a small chemical language model with GA for optimization.  %\cite 

\textbf{Test-time steering of autoregressive language models}
Recently, several approaches have been proposed to steer the output of language models to desired outputs without retraining the entire model. Deng. et al~\cite{deng2023reward} proposed Reward-guided Decoding,  which uses a reward model to score generations as they are produced and rescales sampling probabilities to favor high-reward tokens. Another approach is Self-disciplined Autoregressive Sampling (SASA)~\cite{ko2024large}, which uses the contextual representations learned from the LLM itself to guide it to generate non-toxic text. Lee et. al~\cite{lee2024programming}, uses conditional activation steering to selectively apply or withhold activation steering using LLM activation patterns. While \textsc{GP-MoLFormer-Sim} also relies on test-time steering of a (chemical) language model, different from the prior works it does not involve training of an external or an internal reward model to be used as guidance during decoding, nor does it require analyzing activation patterns of the decoder. Rather, the proposed method exploits contextual similarity with the target at each step of decoding and updates the logits accordingly. 

%\section{Model and Method Details} \PD{Add some details of GP-MoLFormer, hyperparameter used during GG, hyperparameter used during GA. }

\section{Experiments}\label{experiments}
\subsection{Similarity-guided molecule generation}
This task involves generating chemical SMILES similar to a query molecule. We consider five trypsin inhibitors from \cite{hartenfeller2012dogs} as the targets. The baselines considered are random sampling from a 50k pool of unconditionally generated molecules using \textsc{GP-MoLFormer}, a random search in the reaction template and reactant space until a termination condition is met \cite{gottipati2020learningnavigatesyntheticallyaccessible}, and a RL-tuned graph isomorphism network (GIN) model that rewards molecules of high similarity to the target \cite{guptasimilarity}. We report mean Tanimoto similarity ($T_{sim}$), estimated using Morgan fingerprints with a radius of 2, as well as min, max, and mean drug-likeness (QED) over 5 generated sets (one per target) containing the top-k most similar molecules ($k= 1, 10, 10^2, 10^3$ and $10^4$).

\begin{table}[t]
\scriptsize
\centering
\caption{Average similarity and QED values of the five trypsin inhibitor targeted generations, generated by various methods.}
\label{tab:trypsin}
\begin{tabular}{lcHHcHcccHH|lcHHcHcccHH}
\toprule
       &       & SA   & \multicolumn{3}{c}{$T_{sim}$} & \multicolumn{3}{c}{QED} & Guide-QED & Guide-SA &        &       & SA   & \multicolumn{3}{c}{$T_{sim}$} & \multicolumn{3}{c}{QED} & Guide-QED & Guide-SA \\
\cmidrule(lr){5-5}\cmidrule(lr){7-9}\cmidrule(lr){16-16}\cmidrule(lr){18-20}
Config & top-k & mean & min & mean & max              & min & mean & max        & mean      & mean     & Config & top-k & mean & min & mean & max              & min & mean & max        & mean      & mean \\ 
% Config &  Top-k &  Mean &  Min &  Mean &  Max &  QED &  QED &  QED  &  Guide-QED &  Mean &  Config &  Top-k &  Mean &  Min &  Mean &  Max &  QED &  QED &  QED  &  Guide-QED &  Mean\\
%        &        &  SA  &  TSim &  Sim &  TSim &  Min &  Mean &  Max &  Mean   &  Guide-SA  &        &  SA  &  TSim &  Sim &  TSim &  Min &  Mean &  Max &  Mean   &  Guide-SA\\
\midrule
 &      $10^0$ &    3.650 &     1.000 &      1.000 &     1.000 &    0.225 &     0.225 &    0.225 &            &      &  &    $10^0$ &  & & 0.694 & & 0.289 & 0.289 & 0.289 & &      \\
 &     $10^1$ &    3.167 &     0.909 &      0.972 &     1.000 &    0.158 &     0.238 &    0.357 &            &    &   &  $10^1$ &    & & 0.618 & & 0.12 & 0.201 & 0.289 & &        \\
GPMFS &    $10^2$ &    3.192 &     0.826 &      0.877 &     1.000 &    0.075 &     0.241 &    0.555 &           0.234 &          2.905 &   S Model\cite{guptasimilarity} &    $10^2$ & & & 0.554 & & 0.049 & 0.206 & 0.712 & &\\
(Ours) &   $10^3$ &    3.347 &     0.704 &      0.763 &     1.000 &    0.037 &     0.240 &    0.738 &           &    &   &   $10^3$ &      & & 0.499 & & 0.024 & 0.262 & 0.923 & &      \\
&  $10^4$ &    3.402 &     0.449 &      0.573 &     1.000 &    0.016 &     0.213 &    0.901 &     &       &      &  $10^4$ & & & 0.439 & & 0.013 & 0.302 & 0.946 & &     \\
\hline
      &      $10^0$ &    2.135 &     0.438 &      0.438 &     0.438 &    0.555 &     0.555 &    0.555 &            &   &  &      $10^0$ &   &&0.477 && 0.483 & 0.483 & 0.483 &&       \\
      
      Random&     $10^1$ &    2.555 &     0.371 &      0.391 &     0.438 &    0.141 &     0.491 &    0.722 &        &    &     Random &     $10^1$ & &&   0.45 && 0.353 & 0.448 & 0.559 &&       \\
     Generations   &    $10^2$ &    2.684 &     0.326 &      0.348 &     0.438 &    0.048 &     0.520 &    0.866 &           n/a &          2.905 & Search\cite{guptasimilarity} & $10^4$ & &&  0.417 && 0.109 & 0.418 & 0.841 &&  \\
      &   $10^3$ &    2.672 &     0.264 &      0.290 &     0.438 &    0.019 &     0.550 &    0.943 &            &   &  &   $10^3$ &  && 0.377 && 0.041 & 0.385 & 0.841 &&    \\
      &  $10^4$ &    2.830 &     0.193 &      0.225 &     0.438 &    0.011 &     0.571 &    0.947 &            &     &  &  $10^4$ &   && 0.333 && 0.022 & 0.316 & 0.929 &&      \\
\bottomrule
\end{tabular}
\end{table}

Table \ref{tab:trypsin} reports the mean similarity ($T_{sim}$) of top-k most similar generations obtained using \textsc{GP-MoLFormer-Sim} (GPMFS) and baseline methods. Similarity values of the top-ranked generations show that target similarity-guided decoding using GPMFS performs better than test-time baselines like random sampling and random search. The proposed method also outperforms a graph generative model that is RL-tuned to optimize the target similarity (S model) across all values of k up to $10^4$. The reported min, mean, and max QED values show the inverse relation between QED and similarity for these targets, given the mean QED value of these five targets is only 0.234. Nevertheless,  132 molecules are found to have  a QED value $>0.7$ in the  top-10000 most similar molecules generated by GPMFS. Those show a mean similarity of 0.47 and a max similarity of 0.67, showcasing the potential of the proposed method to guide generations on-the-fly toward a target molecule, while yielding useful molecules. A small sample of molecules generated by the GPMFS and their respective targets are visualized in Figure \ref{fig:trypsin_clusters} for varying parameter settings (final parameter values can be found in the Appendix Table \ref{tab:hyperparameters}).

\subsection{Sample-efficient molecular optimization --- PMO benchmark}
The open-source benchmark for practical molecular optimization, PMO \cite{gao2022sample}, has served as an  enabler for the transparent and robust evaluation of diverse sets of molecular optimization algorithms. It involves 23 single-objective  optimization tasks, that includes  property optimization, molecular rediscovery, and structure-based drug design,  with a specific focus on the sample efficiency. PMO includes comparing optimization algorithms involving  reinforcement learning, Bayesian optimization, generative models, GFlowNets, and genetic algorithms. 
We compare \textsc{GP-MoLFormer-Sim+GA} (GPMFS+GA) with existing GA-based molecular optimization methods on this benchmark, while focusing on sample efficiency. Following Gao, et al.~\cite{gao2022sample}, we measure the performance by the area under the curve (AUC) of the average property scores of the top-10 molecules versus oracle calls, with the number of maximum oracle calls being 10k. We utilize the task-specific oracles implemented in the Therapeutics Data Commons (TDC) library \cite{huang2021therapeuticsdatacommonsmachine}. Average and standard deviation of scores obtained from five independent runs starting from different random seeds are reported, unless stated otherwise. 

\begin{table}[t]
\scriptsize
\centering
\caption{Comparison of the guided generation \textsc{GP-MoLFormer-Sim+GA} (``GPMFS+GA'') to selected training-free GA-based baselines. The values for Graph-GA, STONED SELFIES and SynNet are taken from~\cite{gao2022sample}. Mean ($\pm$ standard deviation) AUC top-10 over 5 runs for each.}
\begin{tabular}{lcccccccc}
\toprule
                    Task &  Our Rank & GPMFS+GA & Graph-GA  & STONED  & SynNet &        MOL-GA & MOLLEO  \\
                         &    & (Ours) &  \cite{jensen2019graph}  & SELFIES \cite{Nigam_Pollice_Krenn_Gomes_Aspuru-Guzik_2021}&  Synthesis \cite{gao2021amortized}&      \cite{tripp2023geneticalgorithmsstrongbaselines}   & (GPT-4)\cite{wang2024efficient} \\
\midrule
    albuterol\_similarity &         4 &              0.824 (.071) &      0.838 (.016) &  0.745 (.076) &    0.584 (.039) & 0.896 (.035) &  0.985 (.024) \\
          amlodipine\_mpo &         3 &              0.680 (.064) &      0.661 (.020) &  0.608 (.046) &    0.565 (.007) & 0.688 (.039) &  0.773 (.037) \\
   celecoxib\_rediscovery &         2 &              0.716 (.067) &      0.630 (.097) &  0.382 (.041) &    0.441 (.027) & 0.567 (.083) &  0.864 (.034) \\
                deco\_hop &         2 &              0.710 (.058) &      0.619 (.004) &  0.611 (.008) &    0.613 (.009) & 0.649 (.025) &  0.942 (.013) \\
                    DRD2 &         4 &              0.956 (.010) &      0.964 (.012) &  0.913 (.020) &    0.969 (.004) & 0.936 (.016) &  0.968 (.012) \\
        fexofenadine\_mpo &         3 &              0.798 (.028) &      0.760 (.011) &  0.797 (.016) &    0.761 (.015) & 0.825 (.019) &  0.847 (.018) \\
                    GSK3 &         1 &              0.896 (.035) &      0.788 (.070) &  0.668 (.049) &    0.789 (.032) & 0.843 (.039) &  0.863 (.047) \\
        isomers\_c7h8n2o2 &         2 &              0.932 (.011) &      0.862 (.065) &  0.899 (.011) &    0.455 (.031) & 0.878 (.026) &  0.984 (.008) \\
  isomers\_c9h10n2o2pf2cl &         3 &              0.864 (.016) &      0.719 (.047) &  0.805 (.031) &    0.241 (.064) & 0.865 (.012) &  0.874 (.053) \\
                    JNK3 &         1 &              0.806 (.087) &      0.553 (.136) &  0.523 (.092) &    0.630 (.034) & 0.702 (.123) &  0.790 (.027) \\
                 median1 &         2 &              0.340 (.034) &      0.294 (.021) &  0.266 (.016) &    0.218 (.008) & 0.257 (.009) &  0.352 (.024) \\
                 median2 &         4 &              0.255 (.031) &      0.273 (.009) &  0.245 (.032) &    0.235 (.006) & 0.301 (.021) &  0.275 (.045) \\
    mestranol\_similarity &         2 &              0.658 (.118) &      0.579 (.022) &  0.609 (.101) &    0.399 (.021) & 0.591 (.053) &  0.972 (.009) \\
         osimertinib\_mpo &         5 &              0.819 (.004) &      0.831 (.005) &  0.822 (.012) &    0.796 (.003) & 0.844 (.015) &  0.835 (.024) \\
         perindopril\_mpo &         2 &              0.584 (.042) &      0.538 (.009) &  0.488 (.011) &    0.557 (.011) & 0.547 (.022) &  0.600 (.031) \\
                     QED &         6 &              0.940 (.001) &      0.940 (.000) &  0.941 (.000) &    0.941 (.000) & 0.941 (.001) &  0.948 (.000) \\
          ranolazine\_mpo &         1 &              0.812 (.024) &      0.728 (.012) &  0.765 (.029) &    0.741 (.010) & 0.804 (.011) &  0.769 (.022) \\
            scaffold\_hop &         2 &              0.531 (.016) &      0.517 (.007) &  0.521 (.034) &    0.502 (.012) & 0.527 (.025) &  0.971 (.004) \\
         sitagliptin\_mpo &         3 &              0.501 (.081) &      0.433 (.075) &  0.393 (.083) &    0.025 (.014) & 0.582 (.040) &  0.584 (.067) \\
 thiothixene\_rediscovery &         3 &              0.504 (.033) &      0.479 (.025) &  0.367 (.027) &    0.401 (.019) & 0.519 (.041) &  0.727 (.052) \\
troglitazone\_rediscovery &         2 &              0.437 (.067) &      0.390 (.016) &  0.320 (.018) &    0.283 (.008) & 0.427 (.031) &  0.562 (.019) \\
        valsartan\_smarts &         2 &              0.158 (.317) &      0.000 (.000) &  0.000 (.000) &    0.000 (.000) & 0.000 (.000) &  0.867 (.092) \\
            zaleplon\_mpo &         3 &              0.504 (.022) &      0.346 (.032) &  0.325 (.027) &    0.341 (.011) & 0.519 (.029) &  0.510 (.031) \\
                 Average & 2.7       &              0.662 (.221) &      0.597 (.233) &  0.566 (.245) &    0.499 (.259) & 0.639 (.236) &  0.777 (.200) \\
      Rank by avg. score &        -- &                          2 &                 4 &             5 &              6 &            3 &            1 \\
\bottomrule
\end{tabular}
\label{tab:pmo-ga}
\end{table}

Table \ref{tab:pmo-ga} reports the performance of GPMFS+GA on the 23 tasks from the PMO benchmark. Since the proposed method relies on a combination of the test-time steering of the deep generative model  and a modified genetic algorithm, in the main article we show comparison of the proposed method with GA-based baselines specifically designed for molecular design that do not require any training of the generative model. The baselines shown in Table \ref{tab:pmo-ga} are Graph GA \cite{jensen2019graph}, STONED \cite{Nigam_Pollice_Krenn_Gomes_Aspuru-Guzik_2021}, SynNet \cite{gao2021amortized}, Mol-GA \cite{tripp2023geneticalgorithmsstrongbaselines}, and MOLLEO \cite{wang2024efficient}.  We report the rank per task based on the top-10 AUC score obtained with a maximum of 10k oracle calls for each method.  Average rank and average score over all tasks are also reported in Table \ref{tab:pmo-ga}. For comparison with additional baselines and more analyses on the optimization runs, see Appendix. Results show that the proposed method scores second among GA baselines, while MOLLEO that uses Graph-GA with GPT-4 comes first. On three tasks, namely GSK3, JNK3, and ranolazine\_mpo, GPMFS+GA outperforms all baselines, while on another 9 tasks it ranks second.   Given the computational and actual dollar cost associated with calling GPT-4, GPMFS+GA appears as a more cost-effective alternative. %\PD{optimization plot. Attn Jiri.}\jn{see Fig. \ref{fig:optimization_curves}, also Figure \ref{fig:albuterol_opt_traj} }
To gauge synthesizability, average Synthetic Accessibility scores of the top 100 generated molecules in each task are reported in Appendix Table \ref{tab:sa}.

We also compare GPMFS+GA  with other MOLLEO variants that use a smaller domain-aware language model --- namely, a BioT5 model and  a MoleculeSTM model (see Appendix Table \ref{tab:bsl_comparison_wang}). GPMFS+GA performs better than MOLLEO (MoleculeSTM), while losing against the BioT5 variant. 

\begin{table}[t]
    \caption{Comparison of \textsc{GP-MoLFormer-Sim+GA} with MOLLEO in the black-box oracle setting. Molecule name in prompt is redacted for MOLLEO in that setting.  Mean ($\pm$ standard deviation) AUC top-10 over 5 runs for each.}

    \centering
    \begin{tabular}{lccc}
        \toprule
         & \multicolumn{2}{c}{black-box} & \\
           \cmidrule(lr){2-3}
        Task & GPMFS+GA & MOLLEO & MOLLEO \\
         & (ours) & (GPT-4.1-mini, redacted) &  (GPT-4.1-mini) \\ \midrule
        thiothixene\_rediscovery & 0.504 (.033) & 0.462 (.031) & 0.692 (.013) \\
        mestranol\_similarity & 0.658 (.118) & 0.644 (.065) & 0.983 (.001) \\
        \bottomrule
    \end{tabular}
    \label{tab:redacted}
\end{table}

\subsection{Comparison with MOLLEO in the black-box oracle setting}
\label{sec:black-box_results}
MOLLEO includes natural language prompting of the LLM to generate proposals based on the GA operations --- crossover and mutation. While doing so, the prompt includes information about the  task and the oracle function. For example, for the thiothixene rediscovery task, the prompt used in reference \cite{wang2024efficient} includes the following (\emph{emphasis} ours): ``OBJECTIVE: has a higher \emph{thiothixene} rediscovery score.  TASK: \emph{thiothixene} rediscovery scores. OBJECTIVE\_DEFINITION: The \emph{thiothixene} rediscovery score measures a molecule’s Tanimoto
similarity with \emph{thiothixene’s} SMILES to check whether it could be rediscovered.'' This contextual information present in the prompt weakens the black-box nature of the oracle used in the PMO benchmark as the GPT-4 model has memorized the thiothixene SMILES, which is otherwise never disclosed to the other baselines. Therefore, to enable a fair comparison, we revise the prompt such that the name of the target molecule is redacted from the prompt. Results are reported in Table \ref{tab:redacted} for two exemplar tasks, namely thiothixene rediscovery and mestranol similarity. On both tasks, MOLLEO's performance (using a {\tt gpt-4.1-mini} model) drops by $\sim$33\% and becomes worse compared to GPMFS+GA %shows a performance drop from  0.692+/-0.013 (using a  gpt-4.1-mini model), when the original prompt is used, to 0.462+/-0.031, 
when the molecule name is redacted. %In comparison, GPMFS+GA yields a score of 0.504+/-0.022.  
This result implies that MOLLEO's performance depends on the LLM's utilization of the task-relevant contextual information for proposing offspring. In contrast, our proposed method only uses the  score from the (black-box) oracle, consistent with the setting of the PMO benchmark, to produce candidates during optimization and outperforms MOLLEO in that mode.

We also run an experiment where GPMFS+GA has access to the target SMILES information used in the oracle function (when applicable) and utilizes that to create the initial pool of candidates for optimization. Table \ref{tab:target-seeded-metrics} shows the performance gain achieved by the proposed method in that mode, again underscoring the inflationary effect of breaking the black-box nature of the oracle on the performance.

\begin{table}[t]
\centering
\footnotesize
\caption{Top-10 AUC metrics for \textsc{GP-MoLFormer-Sim+GA} (``GPMFS+GA'') under access to the oracle's target SMILES in comparison to MOLLEO models. Mean ($\pm$ standard deviation) AUC top-10 over 5 runs for each.}
\begin{tabular}{lccccc}
\toprule
                    Task &  Our  & GPMFS+GA & MOLLEO & MOLLEO & MOLLEO\cite{wang2024efficient} \\
                         &  Rank & (Ours) & (MolSTM) & (BioT5) & (GPT-4) \\
\midrule
    albuterol&         1 &             0.994 (.005) &   0.929 (.005) &  0.968 (.003) &  0.985 (.024) \\
          amlodipine&         3 &             0.719 (.061) &   0.674 (.018) &  0.776 (.038) &  0.773 (.037) \\
   celecoxib&         1 &             0.885 (.010) &   0.594 (.105) &  0.508 (.017) &  0.864 (.034) \\
        fexofenadine&         2 &             0.812 (.046) &   0.789 (.016) &  0.773 (.017) &  0.847 (.018) \\
                 median1 &         1 &             0.405 (.001) &   0.298 (.019) &  0.338 (.033) &  0.352 (.024) \\
                 median2 &         1 &             0.393 (.004) &   0.251 (.031) &  0.259 (.019) &  0.275 (.045) \\
    mestranol&         1 &             0.993 (.011) &   0.596 (.018) &  0.717 (.104) &  0.972 (.009) \\
         osimertinib&         3 &             0.818 (.002) &   0.823 (.007) &  0.817 (.016) &  0.835 (.024) \\
         perindopril&         3 &             0.582 (.018) &   0.554 (.037) &  0.738 (.016) &  0.600 (.031) \\
          ranolazine&         1 &             0.825 (.016) &   0.725 (.040) &  0.749 (.012) &  0.769 (.022) \\
         sitagliptin&         4 &             0.435 (.054) &   0.548 (.065) &  0.506 (.100) &  0.584 (.067) \\
 thiothixene&         1 &             0.862 (.020) &   0.508 (.035) &  0.696 (.081) &  0.727 (.052) \\
troglitazone&         1 &             0.905 (.002) &   0.381 (.025) &  0.390 (.044) &  0.562 (.019) \\
            zaleplon&         1 &             0.684 (.024) &   0.475 (.018) &  0.465 (.026) &  0.510 (.031) \\
                 Average & 1.7 &             0.737 (.201) &   0.582 (.189) &  0.621 (.201) &  0.690 (.209) \\
      Rank by avg. score &        -- &                         1 &               4 &              3 &              2 \\
\bottomrule
\end{tabular}
\label{tab:target-seeded-metrics}
\end{table}

\subsection{Ablation experiments}

Algorithm \ref{alg:GA} in Section \ref{sec:gpmfs-ga} provides for several optional steps, namely: (1) using random Fourier features (RFF) to approximate a kernel distance in the \textsc{GP-MoLFormer} embedding space, (2) using a genetic graph-based crossover (XO) between parent molecules (drawn from the set of best guides), and (3) adding a set of diverse guides (DIV) to enhance exploration. Results reported in Tables \ref{tab:pmo-ga} and \ref{tab:target-seeded-metrics} were obtained employing all of these variants active. To tease apart individual effects of these options, we also ran a series of ablation experiments over the 23 PMO tasks. Table \ref{tab:summary_ablation_mpo} summarizes the relative performance in multiple configurations, starting with guided generation (GG) with none of the three options to GG with the full set active. For each combination we report the average rank over all tasks as well as the average AUC metric. We observe each option adding a benefit, with the exception of the RFF, however, only when being added alone. The best configuration is {\tt GG+RFF768+XO+DIV} which is used throughout our PMO experiments, unless otherwise stated. Detailed, per-task ablation results are given in the Appendix (Table \ref{Tab:PMOAblation}) along with further hyperparameter details in Table \ref{tab:hyperparameters}.

\begin{table}[t]
\centering
\caption{Overall improvement due to specific method combinations in terms of average 1-based rank and top-10 AUC metric in \textsc{GP-MoLFormer-Sim}. Includes adding ``XO'' as crossover, ``RFF768'' as 768-dimensional Random Fourier Features, and ``DIV'' as diversity augmentation to the vanilla Guided Generation (GG).}
\begin{tabular}{lcc}
\toprule
        Config &  Avg. rank $\downarrow$ &  Avg. score $\uparrow$\\
\midrule
            GG &      5.0 &       0.603 \\
           +XO &      3.4 &       0.672 \\
       +RFF768 &      5.3 &       0.597 \\
       +XO+DIV &      3.0 &       0.678 \\
    +RFF768+XO &      2.5 &       0.682 \\
+RFF768+XO+DIV &      1.8 &       0.690 \\
\bottomrule
\end{tabular}
\label{tab:summary_ablation_mpo}
\end{table}

\section{Limitations and Broader Impact} \label{limitations}
Given the need for discovering new and useful artifacts for various discovery applications, the proposed method can have broader impact beyond  chemistry and biology.
There remain open questions and limitations, however. For example, although the proposed framework is model and domain-agnostic, we have only experimented here with a specific chemical language model decoder and on specific molecular optimization tasks. It also remains an open question to what extent the method can benefit from including ``negative'' targets while subjected to an optimization task.  Extending the method to multi-objective optimization (optimizing linear combination of multiple objectives with different importances) can also be a potential future research direction.

\section{Conclusions}
In this work, we present \textsc{GP-MoLFormer-Sim}, a test-time framework for sequentially  revising the generated output of a small chemical language model to maintain the contextual closeness between its generation and a given set of targets. Furthermore, we integrate the proposed method into a  genetic algorithm as an effective proposer of mutations to produce high-quality offspring (\textsc{GP-MoLFormer-Sim+GA}).  Our method is validated on a variety of molecular optimization tasks. Evaluating this framework with a black-box oracle reveals performance improvement compared to a baseline that leverages large language model like GPT-4, demonstrating the effective trade-off between performance and computational efficiency of the proposed method. We believe the proposed guided generation method represents a versatile and valuable addition to the modeling toolbox in molecular optimization and beyond.

\bibliographystyle{plain}  % or try: alpha, abbrv, ieeetr, etc.
\bibliography{confs}

@string{ICLR = {{Proc. of the International Conference on Learning Representations}}}

@article{gomez2018automatic,
  title={Automatic chemical design using a data-driven continuous representation of molecules},
  author={G{\'o}mez-Bombarelli, Rafael and Wei, Jennifer N and Duvenaud, David and Hern{\'a}ndez-Lobato, Jos{\'e} Miguel and S{\'a}nchez-Lengeling, Benjam{\'\i}n and Sheberla, Dennis and Aguilera-Iparraguirre, Jorge and Hirzel, Timothy D and Adams, Ryan P and Aspuru-Guzik, Al{\'a}n},
  journal={ACS Central Science},
  volume={4},
  number={2},
  pages={268--276},
  year={2018},
  publisher={ACS Publications}
}

@article{lee2023drug,
  title={Drug discovery with dynamic goal-aware fragments},
  author={Lee, Seul and Lee, Seanie and Kawaguchi, Kenji and Hwang, Sung Ju},
  journal={arXiv preprint arXiv:2310.00841},
  year={2023}
}

@inproceedings{lee2023exploring,
  title={Exploring chemical space with score-based out-of-distribution generation},
  author={Lee, Seul and Jo, Jaehyeong and Hwang, Sung Ju},
  booktitle={International Conference on Machine Learning},
  pages={18872--18892},
  year={2023},
  organization={PMLR}
}

@article{wang2024efficient,
  title={Efficient evolutionary search over chemical space with large language models},
  author={Wang, Haorui and Skreta, Marta and Ser, Cher-Tian and Gao, Wenhao and Kong, Lingkai and Strieth-Kalthoff, Felix and Duan, Chenru and Zhuang, Yuchen and Yu, Yue and Zhu, Yanqiao and others},
  journal={arXiv preprint arXiv:2406.16976},
  year={2024}
}

@inproceedings{
    neil2018exploring,
    title={Exploring deep recurrent models with reinforcement learning for molecule design},
    author={Neil, Daniel and Segler, Marwin and Guasch, Laura and Ahmed, Mohamed and Plumbley, Dean and Sellwood, Matthew and Brown, Nathan},
    booktitle={ICLR},
    year={2019},
   
}

@article{bengio2021flow,
  title={Flow network based generative models for non-iterative diverse candidate generation},
  author={Bengio, Emmanuel and Jain, Moksh and Korablyov, Maksym and Precup, Doina and Bengio, Yoshua},
  journal={Advances in Neural Information Processing Systems},
  volume={34},
  pages={27381--27394},
  year={2021}
}

@article{bengio2023gflownet,
  title={Gflownet foundations},
  author={Bengio, Yoshua and Lahlou, Salem and Deleu, Tristan and Hu, Edward J and Tiwari, Mo and Bengio, Emmanuel},
  journal={Journal of Machine Learning Research},
  volume={24},
  number={210},
  pages={1--55},
  year={2023}
}

@inproceedings{rff,
 author = {Rahimi, Ali and Recht, Benjamin},
 booktitle = {Advances in Neural Information Processing Systems},
 editor = {J. Platt and D. Koller and Y. Singer and S. Roweis},
 pages = {},
 publisher = {Curran Associates, Inc.},
 title = {Random Features for Large-Scale Kernel Machines},
 url = {https://proceedings.neurips.cc/paper_files/paper/2007/file/013a006f03dbc5392effeb8f18fda755-Paper.pdf},
 volume = {20},
 year = {2007}
}

@inproceedings{shen2023tacogfn,
  title={Tacogfn: Target conditioned gflownet for drug design},
  author={Shen, Tony and Pandey, Mohit and Ester, Martin},
  booktitle={NeurIPS 2023 Generative AI and Biology (GenBio) Workshop},
  year={2023}
}

@article{loeffler2024reinvent,
  title={Reinvent 4: Modern AI--driven generative molecule design},
  author={Loeffler, Hannes H and He, Jiazhen and Tibo, Alessandro and Janet, Jon Paul and Voronov, Alexey and Mervin, Lewis H and Engkvist, Ola},
  journal={Journal of Cheminformatics},
  volume={16},
  number={1},
  pages={20},
  year={2024},
  publisher={Springer}
}

@article{weininger1988smiles,
  title={SMILES, a chemical language and information system. 1. Introduction to methodology and encoding rules},
  author={Weininger, David},
  journal={Journal of chemical information and computer sciences},
  volume={28},
  number={1},
  pages={31--36},
  year={1988},
  publisher={ACS Publications}
}

@article{olivecrona2017molecular,
  title={Molecular de-novo design through deep reinforcement learning},
  author={Olivecrona, Marcus and Blaschke, Thomas and Engkvist, Ola and Chen, Hongming},
  journal={Journal of Cheminformatics},
  volume={9},
  number={1},
  pages={48},
  year={2017},
  publisher={Nature Publishing Group}
}

@article{zhou2019optimization,
  title={Optimization of molecules via deep reinforcement learning},
  author={Zhou, Zhenpeng and Kearnes, Steven and Li, Li and Zare, Richard N and Riley, Patrick},
  journal={Scientific Reports},
  volume={9},
  number={1},
  pages={10752},
  year={2019},
  publisher={Nature Publishing Group}
}

@article{lee2024programming,
  title={Programming refusal with conditional activation steering},
  author={Lee, Bruce W and Padhi, Inkit and Ramamurthy, Karthikeyan Natesan and Miehling, Erik and Dognin, Pierre and Nagireddy, Manish and Dhurandhar, Amit},
  journal={arXiv preprint arXiv:2409.05907},
  year={2024}
}

@article{ko2024large,
  title={Large Language Models can be Strong Self-Detoxifiers},
  author={Ko, Ching-Yun and Chen, Pin-Yu and Das, Payel and Mroueh, Youssef and Dan, Soham and Kollias, Georgios and Chaudhury, Subhajit and Pedapati, Tejaswini and Daniel, Luca},
  journal={arXiv preprint arXiv:2410.03818},
  year={2024}
}

@article{deng2023reward,
  title={Reward-augmented decoding: Efficient controlled text generation with a unidirectional reward model},
  author={Deng, Haikang and Raffel, Colin},
  journal={arXiv preprint arXiv:2310.09520},
  year={2023}
}

@article{moss2020boss,
  title={Boss: Bayesian optimization over string spaces},
  author={Moss, Henry and Leslie, David and Beck, Daniel and Gonzalez, Javier and Rayson, Paul},
  journal={Advances in neural information processing systems},
  volume={33},
  pages={15476--15486},
  year={2020}
}

@article{segler2017generating,
  title={Generating focused molecule libraries for drug discovery with recurrent neural networks},
  author={Segler, Marwin HS and Kogej, Thierry and Tyrchan, Christian and Waller, Mark P},
  journal={ACS Central Science},
  volume={4},
  number={1},
  pages={120--131},
  year={2017},
  publisher={ACS Publications}
}

@article{blaschke2018application,
  title={Application of generative autoencoder in de novo molecular design},
  author={Blaschke, Thomas and Olivecrona, Marcus and Engkvist, Ola and Bajorath, J{\"u}rgen and Chen, Hongming},
  journal={Molecular Informatics},
  volume={37},
  number={1-2},
  pages={1700123},
  year={2018},
  publisher={Wiley Online Library}
}

@article{jin2018junction,
  title={Junction tree variational autoencoder for molecular graph generation},
  author={Jin, Wengong and Barzilay, Regina and Jaakkola, Tommi},
  journal={arXiv:1802.04364},
  year={2018}
}

@article{ye2025drugassist,
  title={Drugassist: A large language model for molecule optimization},
  author={Ye, Geyan and Cai, Xibao and Lai, Houtim and Wang, Xing and Huang, Junhong and Wang, Longyue and Liu, Wei and Zeng, Xiangxiang},
  journal={Briefings in Bioinformatics},
  volume={26},
  number={1},
  pages={bbae693},
  year={2025},
  publisher={Oxford University Press}
}

@article{hoffman2022optimizing,
  title={Optimizing molecules using efficient queries from property evaluations},
  author={Hoffman, Samuel C and Chenthamarakshan, Vijil and Wadhawan, Kahini and Chen, Pin-Yu and Das, Payel},
  journal={Nature Machine Intelligence},
  volume={4},
  number={1},
  pages={21--31},
  year={2022},
  publisher={Nature Publishing Group UK London}
}

@misc{ross2025gpmolformerfoundationmodelmolecular,
      title={GP-MoLFormer: A Foundation Model For Molecular Generation}, 
      author={Jerret Ross and Brian Belgodere and Samuel C. Hoffman and Vijil Chenthamarakshan and Jiri Navratil and Youssef Mroueh and Payel Das},
      year={2025},
      eprint={2405.04912},
      archivePrefix={arXiv},
      primaryClass={q-bio.BM},
      url={https://arxiv.org/abs/2405.04912}, 
}

@article{Nigam_Pollice_Krenn_Gomes_Aspuru-Guzik_2021, title={Beyond generative models: superfast traversal, optimization, novelty, exploration and discovery (STONED) algorithm for molecules using SELFIES}, volume={12}, DOI={10.1039/D1SC00231G}, abstractNote={Inverse design allows the generation of molecules with desirable physical quantities using property optimization. Deep generative models have recently been applied to tackle inverse design, as they possess the ability to optimize molecular properties directly through structure modification using gradients. While the ability to carry out direct property optimizations is promising, the use of generative deep learning models to solve practical problems requires large amounts of data and is very time-consuming. In this work, we propose STONED – a simple and efficient algorithm to perform interpolation and exploration in the chemical space, comparable to deep generative models. STONED bypasses the need for large amounts of data and training times by using string modifications in the SELFIES molecular representation. First, we achieve non-trivial performance on typical benchmarks for generative models without any training. Additionally, we demonstrate applications in high-throughput virtual screening for the design of drugs, photovoltaics, and the construction of chemical paths, allowing for both property and structure-based interpolation in the chemical space. Overall, we anticipate our results to be a stepping stone for developing more sophisticated inverse design models and benchmarking tools, ultimately helping generative models achieve wider adoption.}, number={20}, journal={Chem. Sci.}, publisher={The Royal Society of Chemistry}, author={Nigam, AkshatKumar and Pollice, Robert and Krenn, Mario and Gomes, Gabriel dos Passos and Aspuru-Guzik, Alán}, year={2021}, pages={7079–7090} }

@article{jensen2019graph,
  title={A graph-based genetic algorithm and generative model/Monte Carlo tree search for the exploration of chemical space},
  author={Jensen, Jan H},
  journal={Chemical science},
  volume={10},
  number={12},
  pages={3567--3572},
  year={2019},
  publisher={Royal Society of Chemistry}
}

@article{Krenn_2020_selfies,
	doi = {10.1088/2632-2153/aba947},
	year = 2020,
	month = {nov},
	publisher = {{IOP} Publishing},
	volume = {1},
	number = {4},
	pages = {045024},
	author = {Mario Krenn and Florian Häse and AkshatKumar Nigam and Pascal Friederich and Alan Aspuru-Guzik},
	title = {Self-referencing embedded strings ({SELFIES}): A 100{\%} robust molecular string representation},
	journal = {Machine Learning: Science and Technology},
}

@inproceedings{
gao2022sample,
title={Sample Efficiency Matters: A Benchmark for Practical Molecular Optimization},
author={Wenhao Gao and Tianfan Fu and Jimeng Sun and Connor W. Coley},
booktitle={Thirty-sixth Conference on Neural Information Processing Systems Datasets and Benchmarks Track},
year={2022},
url={https://openreview.net/forum?id=yCZRdI0Y7G}
}

@article{ahn2020guiding,
  title={Guiding deep molecular optimization with genetic exploration},
  author={Ahn, Sungsoo and Kim, Junsu and Lee, Hankook and Shin, Jinwoo},
  journal={Advances in neural information processing systems},
  volume={33},
  pages={12008--12021},
  year={2020}
}

@misc{kim2024geneticguidedgflownetssampleefficient,
      title={Genetic-guided GFlowNets for Sample Efficient Molecular Optimization}, 
      author={Hyeonah Kim and Minsu Kim and Sanghyeok Choi and Jinkyoo Park},
      year={2024},
      eprint={2402.05961},
      archivePrefix={arXiv},
      primaryClass={q-bio.BM},
      url={https://arxiv.org/abs/2402.05961}, 
}

@misc{gottipati2020learningnavigatesyntheticallyaccessible,
      title={Learning To Navigate The Synthetically Accessible Chemical Space Using Reinforcement Learning}, 
      author={Sai Krishna Gottipati and Boris Sattarov and Sufeng Niu and Yashaswi Pathak and Haoran Wei and Shengchao Liu and Karam M. J. Thomas and Simon Blackburn and Connor W. Coley and Jian Tang and Sarath Chandar and Yoshua Bengio},
      year={2020},
      eprint={2004.12485},
      archivePrefix={arXiv},
      primaryClass={cs.LG},
      url={https://arxiv.org/abs/2004.12485}, 
}

@article{hartenfeller2012dogs,
  title={DOGS: reaction-driven de novo design of bioactive compounds},
  author={Hartenfeller, Markus and Zettl, Heiko and Walter, Miriam and Rupp, Matthias and Reisen, Felix and Proschak, Ewgenij and Weggen, Sascha and Stark, Holger and Schneider, Gisbert},
  journal={PLoS computational biology},
  volume={8},
  number={2},
  pages={e1002380},
  year={2012},
  publisher={Public Library of Science San Francisco, USA}
}

@misc{huang2021therapeuticsdatacommonsmachine,
      title={Therapeutics Data Commons: Machine Learning Datasets and Tasks for Drug Discovery and Development}, 
      author={Kexin Huang and Tianfan Fu and Wenhao Gao and Yue Zhao and Yusuf Roohani and Jure Leskovec and Connor W. Coley and Cao Xiao and Jimeng Sun and Marinka Zitnik},
      year={2021},
      eprint={2102.09548},
      archivePrefix={arXiv},
      primaryClass={cs.LG},
      url={https://arxiv.org/abs/2102.09548}, 
}

@article{guptasimilarity,
  title={A SIMILARITY-AGNOSTIC REINFORCEMENT LEARNING APPROACH FOR LEAD OPTIMIZATION},
  author={Gupta, Abhor and Current, Sean and Ravindran, Balaraman and Batra, Rohit and Raman, Karthik and others},
 url={https://openreview.net/pdf?id=rjLgCkJH79},
journal={openreview},
year={2024}
}

@misc{tripp2023geneticalgorithmsstrongbaselines,
      title={Genetic algorithms are strong baselines for molecule generation}, 
      author={Austin Tripp and José Miguel Hernández-Lobato},
      year={2023},
      eprint={2310.09267},
      archivePrefix={arXiv},
      primaryClass={cs.NE},
      url={https://arxiv.org/abs/2310.09267}, 
}

@inproceedings{tripp2021fresh,
  title={A fresh look at de novo molecular design benchmarks},
  author={Tripp, Austin and Simm, Gregor NC and Hern{\'a}ndez-Lobato, Jos{\'e} Miguel},
  year={2021},
  booktitle={NeurIPS 2021 AI for Science Workshop}
}

@article{gao2021amortized,
  title={Amortized tree generation for bottom-up synthesis planning and synthesizable molecular design},
  author={Gao, Wenhao and Mercado, Roc{\'\i}o and Coley, Connor W},
  journal={arXiv preprint arXiv:2110.06389},
  year={2021}
}

@article{Ertl_Schuffenhauer_2009, title={Estimation of synthetic accessibility score of drug-like molecules based on molecular complexity and fragment contributions}, volume={1}, ISSN={1758-2946}, DOI={10.1186/1758-2946-1-8}, abstractNote={A method to estimate ease of synthesis (synthetic accessibility) of drug-like molecules is needed in many areas of the drug discovery process. The development and validation of such a method that is able to characterize molecule synthetic accessibility as a score between 1 (easy to make) and 10 (very difficult to make) is described in this article.}, number={1}, journal={Journal of Cheminformatics}, author={Ertl, Peter and Schuffenhauer, Ansgar}, year={2009}, month=jun, pages={8}, language={en} }

@misc{peyré2020computationaloptimaltransport,
      title={Computational Optimal Transport}, 
      author={Gabriel Peyré and Marco Cuturi},
      year={2020},
      eprint={1803.00567},
      archivePrefix={arXiv},
      primaryClass={stat.ML},
      url={https://arxiv.org/abs/1803.00567}, 
}

@inproceedings{ko2025large,
title={Large Language Models can Become Strong Self-Detoxifiers},
author={Ching-Yun Ko and Pin-Yu Chen and Payel Das and Youssef Mroueh and Soham Dan and Georgios Kollias and Subhajit Chaudhury and Tejaswini Pedapati and Luca Daniel},
booktitle={The Thirteenth International Conference on Learning Representations},
year={2025},
url={https://openreview.net/forum?id=jY5oml9fe9}
}

\newpage
\appendix

\section{Theoretical Justification for Algorithm \ref{alg:guided_generation}}
\label{sec:theoretical_justification}

Let $V$ be the vocabulary size. Given previously generated tokens represented as context $c$, let $\pi^{\text{target}}_{\mathsf{KDE}}(\cdot|c)$ be the Kernel Density Estimator (KDE) of the target molecules defined  in the embedding space of \textsc{GP-MoLFormer} noted below as $\pi_{\mathrm{ref}}$.  Our algorithm can be seen at each token generation as the closed form  solution of the following optimization problem:
\begin{align*}
\mathsf{P:} \max_{p \in \Delta_{V}} &(1-\alpha)\underbrace{\sum_{i=1}^V p_i \pi^{\text{target}}_{\mathrm{KDE}}(x_i|c)}_{ \text{Similarity to target }\textsf{KDE}} - \underbrace{\mathsf{KL} (p || \pi_{\mathrm{ref}}^{\alpha}(\cdot|c))}_{\mathsf{KL} \textsf{ to \textsc{GP-MoLFormer}}}\\
    &\mathsf{s.t.}~\Delta_{V}= \left\{ p\in [0,1]^{V} \Big| \sum_{i=1}^V p_i =1 \right\},
\end{align*}
where for   $p \in \Delta_{V}$
and  $q \in \mathbb{R}_+^V,$ we define the generalized KL divergence \citep{peyré2020computationaloptimaltransport} as:
 
\[
\mathrm{KL}(p \,\|\, q) = \sum_{i=1}^V p_i \log\left( \frac{p_i}{q_i} \right) + \sum_{i=1}^V q_i - 1.
\]
The optimization problems $\mathsf{P}$ balances between closeness to \textsc{GP-MoLFormer} distribution smoothed by $\alpha$ ($\pi_{\mathrm{ref}}^{\alpha}$) and the similarity to the target molecules kernel density estimator ($\pi^{\text{target}}_{\mathrm{KDE}}$). We use the notion of generalized divergence since $\pi_{\mathrm{ref}}^{\alpha}$ is a smoothed distribution that is no longer normalized. 
Indeed the optimal solution of this problem \citep{ko2025large} is proportional to $ \pi_{\mathrm{ref}}^{\alpha} \exp((1-\alpha)\pi^{\text{target}}_{\mathrm{KDE}} ) = \exp ((1-\alpha)\pi^{\text{target}}_{\mathrm{KDE}} + \alpha \log \pi_{\mathrm{ref}})$. This justifies the computations on lines 23--24 in Algorithm \ref{alg:guided_generation}, where the new guided probability is obtained using a softmax on the interpolation between the KDE estimator of the target molecules and the logits of \textsc{GP-MoLFormer}.

\section{Extended Results --- PMO Tasks}

\subsection{Optimization trajectory --- an example}

\begin{figure}[!ht]
    \centering
    \includegraphics[width=.7\textwidth]{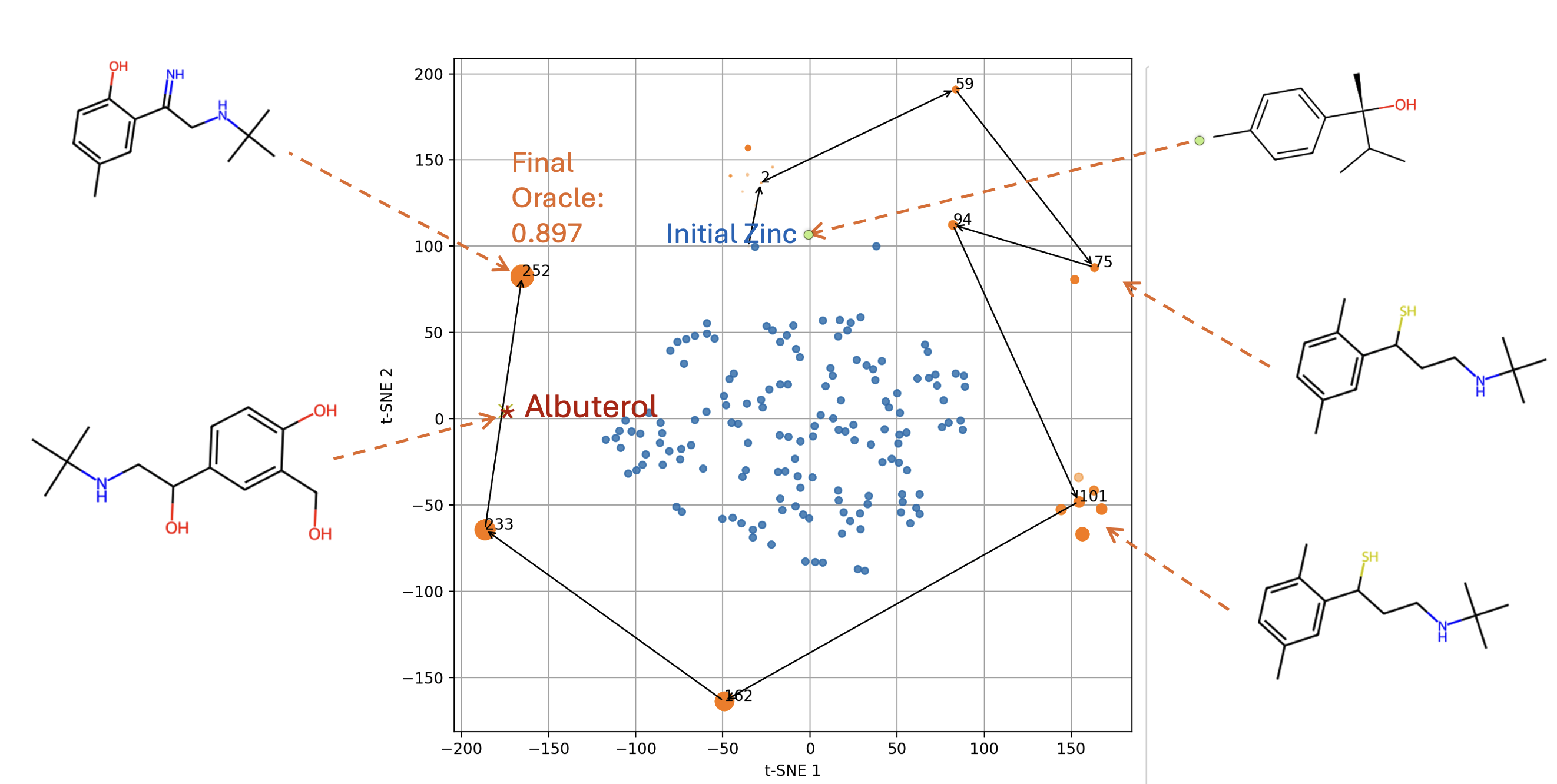}
    \caption{Albuterol optimization trajectory of a single guide visualized in \textsc{GP-MoLFormer} space via a 2D t-SNE projection.} 
    \label{fig:albuterol_opt_traj}
\end{figure}

Figure \ref{fig:albuterol_opt_traj} visualizes the \textsc{GP-MoLFormer-Sim+GA} process in the \textsc{GP-MoLFormer} embedding space (projected onto a 2D t-SNE plane). Starting with a random subset of 100 ZINC molecules (blue dots in the center), the optimization picks up on a handful highest-scoring candidates (molecular structure of one of these is depicted at the top right) and through the guided generation comes up with offspring displayed as dot in orange color (marked as generation 2). The GA process continues a milestones are numbered in Figure \ref{fig:albuterol_opt_traj} by their generation number and shown along with representative chemical structures. The process completes on generation 252 upon exhausting the oracle budget of 10000, reaching a final oracle value of 0.897. Also shown is the actual oracle target, the compound albuterol. Note that only a single guide trajectory is shown in this figure for sake of simplicity. In our albuterol runs reported in Table \ref{tab:pmo-ga} and Figure \ref{fig:optimization_curves}, the actual target was hit exactly. 

\subsection{Redacted MOLLEO Prompts}
Table \ref{tab:redacted_prompts} lists the modified prompts used for gauging performance gain/loss due to knowledge of the oracle target compound (Section \ref{sec:black-box_results}).

\begin{table}[h]
    \caption{Prompts used for redacted MOLLEO. Words in \textit{italics} are modified from the original prompts.}
    \centering
    \scriptsize
    \begin{tabular}{lp{0.45\textwidth}p{0.3\textwidth}}
        \toprule
        Task & Description & Objective \\ \midrule
        thiothixene\_rediscovery & \texttt{I have two molecules and their rediscovery score measures a molecule's Tanimoto similarity with \textit{a particular} SMILES to check whether it could be rediscovered.} & \texttt{Please propose a new molecule that has a higher rediscovery score.} \\
        mestranol\_similarity & \texttt{I have two molecules and their \textit{target} similarity scores. The \textit{target} similarity score measures a molecule's Tanimoto similarity with \textit{a particular target molecule}.} & \texttt{Please propose a new molecule that has a higher \textit{target} similarity score.} \\
        \bottomrule
    \end{tabular}
    
    \label{tab:redacted_prompts}
\end{table}

\subsection{Configuration ablation}

Table \ref{Tab:PMOAblation} gives efficacy details regarding individual features of the \textsc{GPMoLFormer-Sim+GA} procedure, including adding (1) 768-dimensional random Fourier features (``RFF768''), (2) Crossover (``XO''), and Diversity guides (``DIV''). Overall, the full combination {\tt+RFF768+XO+DIV} gives best results, as can also be seen in the aggregate score in the last row of the table. 

\begin{table}[h]
\scriptsize
\centering
\caption{Top-10 AUC PMO metrics obtained using various configurations (mean ($\pm$ standard deviation) over 5 runs for each).}
\label{Tab:PMOAblation}
\begin{tabular}{lcccccc}
\toprule
\toprule
Task & Guided Gen. &            +XO &        +RFF768 &        +XO+DIV &     +RFF768+XO & +RFF768+XO+DIV \\
\midrule
albuterol     &  0.714 (.044) &  0.803 (.045) &  0.767 (.067) &  0.847 (.050) &  0.814 (.039) &  0.824 (.071) \\
amlodipine           &  0.587 (.034) &  0.699 (.036) &  0.584 (.006) &  0.625 (.022) &  0.715 (.067) &  0.680 (.064) \\
amlomestranol &  0.542 (.059) &  0.579 (.072) &  0.542 (.034) &  0.602 (.049) &  0.603 (.054) &  0.658 (.118) \\
celecoxib    &  0.560 (.114) &  0.627 (.052) &  0.577 (.141) &  0.667 (.079) &  0.758 (.070) &  0.716 (.067) \\
deco\_hop                 &  0.748 (.113) &  0.701 (.086) &  0.668 (.088) &  0.649 (.011) &  0.629 (.009) &  0.710 (.058) \\
DRD2                     &  0.929 (.037) &  0.955 (.021) &  0.954 (.026) &  0.958 (.018) &  0.952 (.015) &  0.956 (.010) \\
fexofenadine         &  0.756 (.020) &  0.791 (.012) &  0.801 (.031) &  0.802 (.020) &  0.840 (.030) &  0.798 (.028) \\
GSK3                     &  0.818 (.061) &  0.833 (.073) &  0.781 (.070) &  0.862 (.081) &  0.871 (.066) &  0.896 (.035) \\
isomers\_c7         &  0.916 (.028) &  0.921 (.026) &  0.907 (.035) &  0.912 (.011) &  0.933 (.006) &  0.932 (.011) \\
isomers\_c9   &  0.819 (.025) &  0.842 (.027) &  0.831 (.034) &  0.847 (.023) &  0.861 (.015) &  0.864 (.016) \\
JNK3                     &  0.654 (.037) &  0.773 (.085) &  0.609 (.078) &  0.791 (.038) &  0.801 (.092) &  0.806 (.087) \\
median1                  &  0.277 (.011) &  0.352 (.035) &  0.320 (.032) &  0.358 (.035) &  0.369 (.022) &  0.340 (.034) \\
median2                  &  0.229 (.012) &  0.218 (.012) &  0.204 (.013) &  0.252 (.027) &  0.234 (.028) &  0.255 (.031) \\
osimertimib          &  0.806 (.018) &  0.813 (.011) &  0.807 (.016) &  0.812 (.007) &  0.815 (.009) &  0.819 (.004) \\
perindopril          &  0.531 (.027) &  0.580 (.037) &  0.528 (.033) &  0.551 (.014) &  0.572 (.036) &  0.584 (.042) \\
QED                      &  0.939 (.002) &  0.940 (.001) &  0.939 (.001) &  0.940 (.001) &  0.941 (.000) &  0.940 (.001) \\
ranolazine           &  0.782 (.012) &  0.805 (.009) &  0.799 (.021) &  0.788 (.016) &  0.815 (.007) &  0.812 (.024) \\
scaffold\_hop             &  0.519 (.019) &  0.543 (.018) &  0.511 (.016) &  0.527 (.004) &  0.539 (.012) &  0.531 (.016) \\
sitagliptin          &  0.364 (.073) &  0.451 (.010) &  0.384 (.041) &  0.480 (.032) &  0.445 (.050) &  0.501 (.081) \\
thiothixene  &  0.429 (.008) &  0.526 (.037) &  0.432 (.008) &  0.525 (.012) &  0.461 (.011) &  0.504 (.033) \\
troglitazone &  0.346 (.059) &  0.385 (.090) &  0.329 (.039) &  0.408 (.098) &  0.420 (.051) &  0.437 (.067) \\
valsartan\_smarts         &  0.100 (.199) &  0.040 (.080) &  0.000 (.000) &  0.118 (.180) &  0.073 (.087) &  0.158 (.317) \\
zaleplon             &  0.496 (.029) &  0.493 (.007) &  0.469 (.014) &  0.497 (.017) &  0.487 (.017) &  0.504 (.022) \\
Avg. Score                          &          0.603 &          0.638 &          0.597 &          0.644 &          0.650 &          0.662 \\
\bottomrule
\end{tabular}
\end{table}

\subsubsection{Optimization curves}

\begin{figure}[!ht]
    \centering
    \includegraphics[width=1.\textwidth]{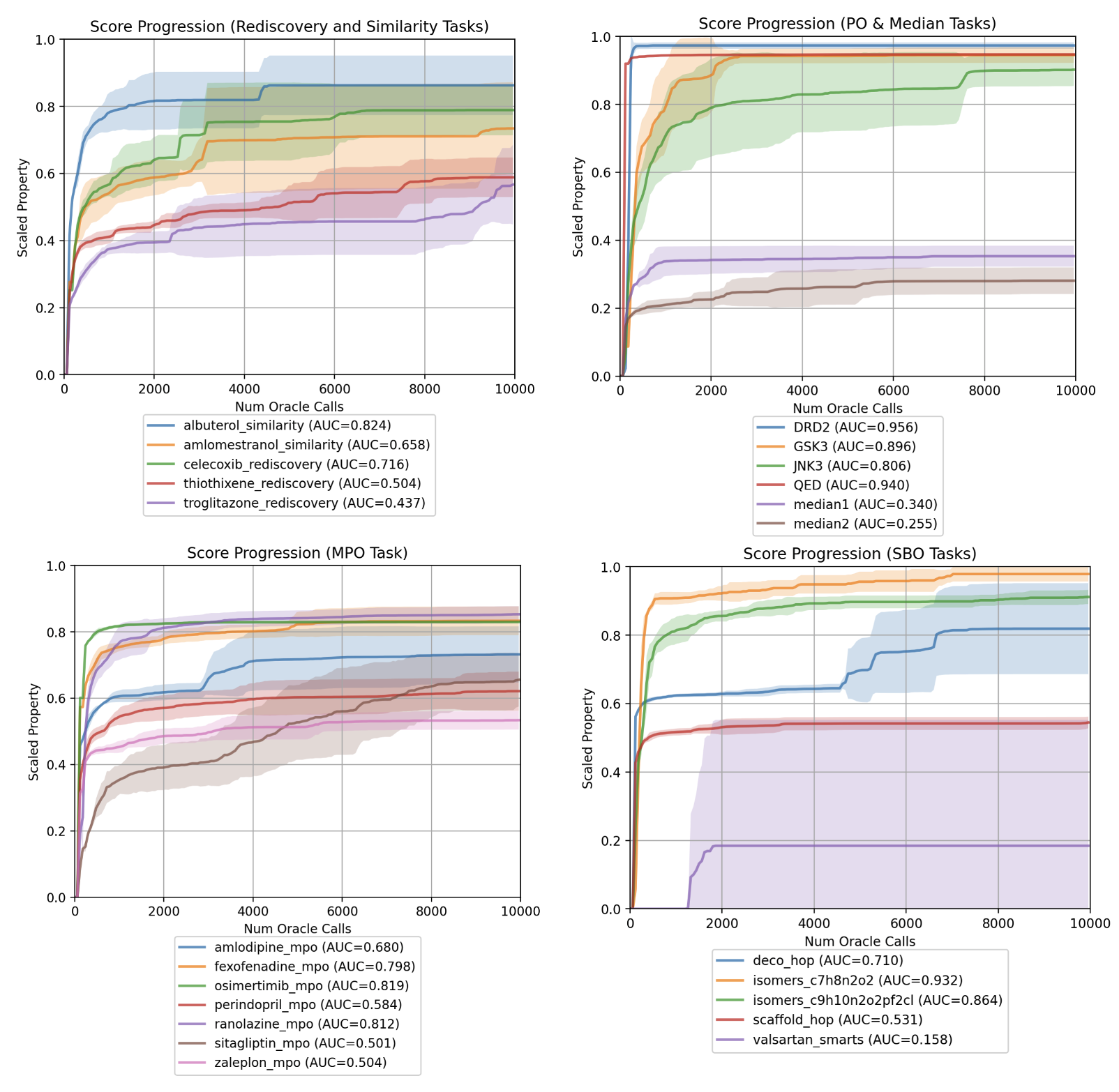}
    \caption{Optimization curves for the PMO tasks by group. } 
    \label{fig:optimization_curves}
\end{figure}

Figure \ref{fig:optimization_curves} shows how the \textsc{GP-MoLFormer-Sim+GA} optimization progresses as a function of number of Oracle calls in the 23 PMO tasks (grouped by type). We observe a variety of patterns: while the similarity, rediscovery and MPO task curves look fairly similar in range and progression, the PO tasks (specifically, QED and DRD2) tend to quickly ramp up with small variance across the 5 runs. On the other hand, the median tasks seem to be significantly more challenging and the process seems to quickly plateau after a few hundred oracle calls. This reflects the nature of the median task and is not unexpected. Finally, the SBO tasks show the largest variability with the special case of valsartan - a task on which most baselines fail to obtain any hits (usually remain flat at 0.0). We believe that our valsartan task result shown in Figure \ref{fig:optimization_curves} highlights the exploratory power of the crossover and diversity enhanced guided generation.  

\subsubsection{Extended baseline comparison}

We list detailed comparison of the \textsc{GP-MoLFormer-Sim GA} top-10 AUC metrics with baselines published previously. 

Table \ref{tab:bsl_comparison_tripp} shows a comparison of our GPMFS-GA to \cite{tripp2023geneticalgorithmsstrongbaselines}. Tables \ref{tab_bsl_comp_gao_p1} through \ref{tab_bsl_comp_gao_p4} show results reported in \cite{gao2022sample}. Tables \ref{tab:bsl_comp_gflow_p1}-\ref{tab:bsl_comp_gflow_p2} compare results from \cite{kim2024geneticguidedgflownetssampleefficient}, and Table \ref{tab:bsl_comparison_wang} compares results published in \cite{wang2024efficient}.

\begin{table}[h]
\centering
\footnotesize
\caption{Comparison of \textsc{GP-MoLFormer-Sim+GA} to main results in \cite{tripp2023geneticalgorithmsstrongbaselines}. Mean ($\pm$ standard deviation) AUC top-10 over 5 runs for each.}
\begin{tabular}{lccccc}
\toprule
                    Task &  Our Rank &       Our AUC &      REINVENT &      Graph GA &        MOL\_GA \\
\midrule
    albuterol&         4 & 0.824 (.071) & 0.882 (.006) & 0.838 (.016) & 0.896 (.035) \\
          amlodipine&         2 & 0.680 (.064) & 0.635 (.035) & 0.661 (.020) & 0.688 (.039) \\
   celecoxib&         1 & 0.716 (.067) & 0.713 (.067) & 0.630 (.097) & 0.567 (.083) \\
                deco\_hop &         1 & 0.710 (.058) & 0.666 (.044) & 0.619 (.004) & 0.649 (.025) \\
                    DRD2 &         2 & 0.956 (.010) & 0.945 (.007) & 0.964 (.012) & 0.936 (.016) \\
        fexofenadine&         2 & 0.798 (.028) & 0.784 (.006) & 0.760 (.011) & 0.825 (.019) \\
                    GSK3 &         1 & 0.896 (.035) & 0.865 (.043) & 0.788 (.070) & 0.843 (.039) \\
        isomers\_c7 &         1 & 0.932 (.011) & 0.852 (.036) & 0.862 (.065) & 0.878 (.026) \\
  isomers\_c9 &         2 & 0.864 (.016) & 0.642 (.054) & 0.719 (.047) & 0.865 (.012) \\
                    JNK3 &         1 & 0.806 (.087) & 0.783 (.023) & 0.553 (.136) & 0.702 (.123) \\
                 median1 &         2 & 0.340 (.034) & 0.356 (.009) & 0.294 (.021) & 0.257 (.009) \\
                 median2 &         4 & 0.255 (.031) & 0.276 (.008) & 0.273 (.009) & 0.301 (.021) \\
    mestranol&         1 & 0.658 (.118) & 0.618 (.048) & 0.579 (.022) & 0.591 (.053) \\
         osimertinib&         4 & 0.819 (.004) & 0.837 (.009) & 0.831 (.005) & 0.844 (.015) \\
         perindopril&         1 & 0.584 (.042) & 0.537 (.016) & 0.538 (.009) & 0.547 (.022) \\
                     QED &         3 & 0.940 (.001) & 0.941 (.000) & 0.940 (.000) & 0.941 (.001) \\
          ranolazine&         1 & 0.812 (.024) & 0.760 (.009) & 0.728 (.012) & 0.804 (.011) \\
            scaffold\_hop &         2 & 0.531 (.016) & 0.560 (.019) & 0.517 (.007) & 0.527 (.025) \\
         sitagliptin&         2 & 0.501 (.081) & 0.021 (.003) & 0.433 (.075) & 0.582 (.040) \\
 thiothixene&         3 & 0.504 (.033) & 0.534 (.013) & 0.479 (.025) & 0.519 (.041) \\
troglitazone&         2 & 0.437 (.067) & 0.441 (.032) & 0.390 (.016) & 0.427 (.031) \\
        valsartan\_smarts &         2 & 0.158 (.317) & 0.178 (.358) & 0.000 (.000) & 0.000 (.000) \\
            zaleplon&         2 & 0.504 (.022) & 0.358 (.062) & 0.346 (.032) & 0.519 (.029) \\
                 Average & 2.0 (1.0) & 0.662 (.221) & 0.617 (.245) & 0.597 (.233) & 0.639 (.236) \\
      Rank by avg. score &        -- &             1 &             3 &             4 &             2 \\
\bottomrule
\end{tabular}
\label{tab:bsl_comparison_tripp}
\end{table}

\begin{table}[h]
\tiny
\centering
\caption{Comparison of \textsc{GP-MoLFormer-Sim+GA} to main results in \cite{gao2022sample} - Page 1 of 4. Mean ($\pm$ standard deviation) AUC top-10 over 5 runs for each.}
\begin{tabular}{lccccccccc}
\toprule
                    Task &   Rank &       Our  & REINVENT & Graph GA & REINVENT  & GP BO & STONED  & LSTM HC & SMILES   \\
                    &   Rank      &       AUC &SMILES & Fragments & SELFIES & Fragments & SELFIES &  SMILES & GA  \\
\midrule
    albuterol&         5 & 0.824 (.071) &   0.882 (.006) &      0.838 (.016) &    0.826 (.030) &   0.898 (.014) &  0.745 (.076) &  0.719 (.018) &    0.661 (.066) \\
          amlodipine&         1 & 0.680 (.064) &   0.635 (.035) &      0.661 (.020) &    0.607 (.014) &   0.583 (.044) &  0.608 (.046) &  0.593 (.016) &    0.549 (.009) \\
   celecoxib&         2 & 0.716 (.067) &   0.713 (.067) &      0.630 (.097) &    0.573 (.043) &   0.723 (.053) &  0.382 (.041) &  0.539 (.018) &    0.344 (.027) \\
                deco\_hop &         4 & 0.710 (.058) &   0.666 (.044) &      0.619 (.004) &    0.631 (.012) &   0.629 (.018) &  0.611 (.008) &  0.826 (.017) &    0.611 (.006) \\
                    DRD2 &         3 & 0.956 (.010) &   0.945 (.007) &      0.964 (.012) &    0.943 (.005) &   0.923 (.017) &  0.913 (.020) &  0.919 (.015) &    0.908 (.019) \\
        fexofenadine&         1 & 0.798 (.028) &   0.784 (.006) &      0.760 (.011) &    0.741 (.002) &   0.722 (.005) &  0.797 (.016) &  0.725 (.003) &    0.721 (.015) \\
                    GSK3 &         1 & 0.896 (.035) &   0.865 (.043) &      0.788 (.070) &    0.780 (.037) &   0.851 (.041) &  0.668 (.049) &  0.839 (.015) &    0.629 (.044) \\
        isomers\_c7 &         1 & 0.932 (.011) &   0.852 (.036) &      0.862 (.065) &    0.849 (.034) &   0.680 (.117) &  0.899 (.011) &  0.485 (.045) &    0.913 (.021) \\
  isomers\_c9 &         1 & 0.864 (.016) &   0.642 (.054) &      0.719 (.047) &    0.733 (.029) &   0.469 (.180) &  0.805 (.031) &  0.342 (.027) &    0.860 (.065) \\
                    JNK3 &         1 & 0.806 (.087) &   0.783 (.023) &      0.553 (.136) &    0.631 (.064) &   0.564 (.155) &  0.523 (.092) &  0.661 (.039) &    0.316 (.022) \\
                 median1 &         3 & 0.340 (.034) &   0.356 (.009) &      0.294 (.021) &    0.355 (.011) &   0.301 (.014) &  0.266 (.016) &  0.255 (.010) &    0.192 (.012) \\
                 median2 &         4 & 0.255 (.031) &   0.276 (.008) &      0.273 (.009) &    0.255 (.005) &   0.297 (.009) &  0.245 (.032) &  0.248 (.008) &    0.198 (.005) \\
    mestranol&         1 & 0.658 (.118) &   0.618 (.048) &      0.579 (.022) &    0.620 (.029) &   0.627 (.089) &  0.609 (.101) &  0.526 (.032) &    0.469 (.029) \\
         osimertinib&         5 & 0.819 (.004) &   0.837 (.009) &      0.831 (.005) &    0.820 (.003) &   0.787 (.006) &  0.822 (.012) &  0.796 (.002) &    0.817 (.011) \\
         perindopril&         1 & 0.584 (.042) &   0.537 (.016) &      0.538 (.009) &    0.517 (.021) &   0.493 (.011) &  0.488 (.011) &  0.489 (.007) &    0.447 (.013) \\
                     QED &         4 & 0.940 (.001) &   0.941 (.000) &      0.940 (.000) &    0.940 (.000) &   0.937 (.000) &  0.941 (.000) &  0.939 (.000) &    0.940 (.000) \\
          ranolazine&         1 & 0.812 (.024) &   0.760 (.009) &      0.728 (.012) &    0.748 (.018) &   0.735 (.013) &  0.765 (.029) &  0.714 (.008) &    0.699 (.026) \\
            scaffold\_hop &         4 & 0.531 (.016) &   0.560 (.019) &      0.517 (.007) &    0.525 (.013) &   0.548 (.019) &  0.521 (.034) &  0.533 (.012) &    0.494 (.011) \\
         sitagliptin&         1 & 0.501 (.081) &   0.021 (.003) &      0.433 (.075) &    0.194 (.121) &   0.186 (.055) &  0.393 (.083) &  0.066 (.019) &    0.363 (.057) \\
 thiothixene&         3 & 0.504 (.033) &   0.534 (.013) &      0.479 (.025) &    0.495 (.040) &   0.559 (.027) &  0.367 (.027) &  0.438 (.008) &    0.315 (.017) \\
troglitazone&         2 & 0.437 (.067) &   0.441 (.032) &      0.390 (.016) &    0.348 (.012) &   0.410 (.015) &  0.320 (.018) &  0.354 (.016) &    0.263 (.024) \\
        valsartan &         2 & 0.158 (.317) &   0.179 (.358) &      0.000 (.000) &    0.000 (.000) &   0.000 (.000) &  0.000 (.000) &  0.000 (.000) &    0.000 (.000) \\
            zaleplon&         1 & 0.504 (.022) &   0.358 (.062) &      0.346 (.032) &    0.333 (.026) &   0.221 (.072) &  0.325 (.027) &  0.206 (.006) &    0.334 (.041) \\
                 Average & 2.3 (1.4) & 0.662 (.221) &   0.617 (.245) &      0.597 (.233) &    0.585 (.242) &   0.571 (.244) &  0.566 (.245) &  0.531 (.258) &    0.524 (.258) \\
      Rank by &&&&&&&&&\\
      avg. score &        -- &             1 &               2 &                  3 &                4 &               5 &              6 &              7 &                8 \\
\bottomrule
\end{tabular}
\label{tab_bsl_comp_gao_p1}
\end{table}

\begin{table}[h]
\tiny
\centering
\caption{Comparison of \textsc{GP-MoLFormer-Sim+GA} to main results in \cite{gao2022sample} - Page 2 of 4. Mean ($\pm$ standard deviation) AUC top-10 over 5 runs for each.}
\begin{tabular}{lccccccc}
\toprule
                    Task & SynNet Synthesis & DoG-Gen Synthesis & DST Fragments &          MARS &        MIMOSA &        MolPal & LSTM HC SELFIES \\
\midrule
    albuterol&    0.584 (.039) &     0.676 (.013) & 0.619 (.020) & 0.597 (.124) & 0.618 (.017) & 0.609 (.002) &   0.664 (.030) \\
          amlodipine&    0.565 (.007) &     0.536 (.003) & 0.516 (.007) & 0.504 (.016) & 0.543 (.003) & 0.582 (.008) &   0.532 (.004) \\
   celecoxib&    0.441 (.027) &     0.464 (.009) & 0.380 (.006) & 0.379 (.060) & 0.393 (.010) & 0.415 (.001) &   0.385 (.008) \\
                deco\_hop &    0.613 (.009) &     0.800 (.007) & 0.608 (.008) & 0.589 (.003) & 0.619 (.003) & 0.643 (.005) &   0.590 (.001) \\
                    DRD2 &    0.969 (.004) &     0.948 (.001) & 0.820 (.014) & 0.891 (.020) & 0.799 (.017) & 0.783 (.009) &   0.729 (.034) \\
        fexofenadine&    0.761 (.015) &     0.695 (.003) & 0.725 (.005) & 0.711 (.006) & 0.706 (.011) & 0.685 (.000) &   0.693 (.004) \\
                    GSK3 &    0.789 (.032) &     0.831 (.021) & 0.671 (.032) & 0.552 (.037) & 0.554 (.042) & 0.555 (.011) &   0.423 (.018) \\
        isomers\_c7 &    0.455 (.031) &     0.465 (.018) & 0.548 (.069) & 0.728 (.027) & 0.564 (.046) & 0.484 (.006) &   0.587 (.031) \\
  isomers\_c9 &    0.241 (.064) &     0.199 (.016) & 0.458 (.063) & 0.581 (.013) & 0.303 (.046) & 0.164 (.003) &   0.352 (.019) \\
                    JNK3 &    0.630 (.034) &     0.595 (.023) & 0.556 (.057) & 0.489 (.095) & 0.360 (.063) & 0.339 (.009) &   0.207 (.013) \\
                 median1 &    0.218 (.008) &     0.217 (.001) & 0.232 (.009) & 0.207 (.011) & 0.243 (.005) & 0.249 (.001) &   0.239 (.009) \\
                 median2 &    0.235 (.006) &     0.212 (.000) & 0.185 (.020) & 0.181 (.011) & 0.214 (.002) & 0.230 (.000) &   0.205 (.005) \\
    mestranol&    0.399 (.021) &     0.437 (.007) & 0.450 (.027) & 0.388 (.026) & 0.438 (.015) & 0.564 (.004) &   0.446 (.009) \\
         osimertinib&    0.796 (.003) &     0.774 (.002) & 0.785 (.004) & 0.777 (.006) & 0.788 (.014) & 0.779 (.000) &   0.780 (.005) \\
         perindopril&    0.557 (.011) &     0.474 (.002) & 0.462 (.008) & 0.462 (.006) & 0.490 (.011) & 0.467 (.002) &   0.448 (.006) \\
                     QED &    0.941 (.000) &     0.934 (.000) & 0.938 (.000) & 0.930 (.003) & 0.939 (.000) & 0.940 (.000) &   0.938 (.000) \\
          ranolazine&    0.741 (.010) &     0.711 (.006) & 0.632 (.054) & 0.740 (.010) & 0.640 (.015) & 0.457 (.005) &   0.614 (.010) \\
            scaffold\_hop &    0.502 (.012) &     0.515 (.005) & 0.497 (.004) & 0.469 (.004) & 0.507 (.015) & 0.494 (.000) &   0.472 (.002) \\
         sitagliptin&    0.025 (.014) &     0.048 (.008) & 0.075 (.032) & 0.016 (.003) & 0.102 (.023) & 0.043 (.001) &   0.116 (.012) \\
 thiothixene&    0.401 (.019) &     0.375 (.004) & 0.366 (.006) & 0.344 (.022) & 0.347 (.018) & 0.339 (.001) &   0.339 (.009) \\
troglitazone&    0.283 (.008) &     0.416 (.019) & 0.279 (.019) & 0.256 (.016) & 0.299 (.009) & 0.268 (.000) &   0.257 (.002) \\
        valsartan\_smarts &    0.000 (.000) &     0.000 (.000) & 0.000 (.000) & 0.000 (.000) & 0.000 (.000) & 0.000 (.000) &   0.000 (.000) \\
            zaleplon&    0.341 (.011) &     0.123 (.016) & 0.176 (.045) & 0.187 (.046) & 0.172 (.036) & 0.168 (.003) &   0.218 (.020) \\
                 Average &    0.499 (.259) &     0.498 (.269) & 0.477 (.236) & 0.477 (.253) & 0.463 (.232) & 0.446 (.238) &   0.445 (.228) \\
      Rank by avg. score &                9 &                10 &            12 &            11 &            13 &            14 &              15 \\
\bottomrule
\end{tabular}
\label{tab_bsl_comp_gao_p2}
\end{table}

\begin{table}[h]
\tiny
\centering
\caption{Comparison of \textsc{GP-MoLFormer-Sim+GA} to main results in \cite{gao2022sample} - Page 3 of 4. Mean ($\pm$ standard deviation) AUC top-10 over 5 runs for each.}
\begin{tabular}{lccccccc}
\toprule
                    Task &        DoG-AE & GFlowNet Fragments &  GA+D SELFIES & VAE BO SELFIES &     Screening & VAE BO SMILES & Pasithea SELFIES \\
\midrule
    albuterol& 0.533 (.034) &      0.447 (.012) & 0.495 (.025) &  0.494 (.012) & 0.483 (.006) & 0.489 (.007) &    0.447 (.007) \\
          amlodipine& 0.507 (.005) &      0.444 (.004) & 0.400 (.032) &  0.516 (.005) & 0.535 (.001) & 0.533 (.009) &    0.504 (.003) \\
   celecoxib& 0.355 (.012) &      0.327 (.004) & 0.223 (.025) &  0.326 (.007) & 0.351 (.005) & 0.354 (.002) &    0.312 (.007) \\
                deco\_hop & 0.765 (.055) &      0.583 (.002) & 0.550 (.005) &  0.579 (.001) & 0.590 (.001) & 0.589 (.001) &    0.579 (.001) \\
                    DRD2 & 0.943 (.009) &      0.590 (.070) & 0.382 (.205) &  0.569 (.039) & 0.545 (.015) & 0.555 (.043) &    0.255 (.040) \\
        fexofenadine& 0.679 (.017) &      0.693 (.006) & 0.587 (.007) &  0.670 (.004) & 0.666 (.004) & 0.671 (.003) &    0.660 (.015) \\
                    GSK3 & 0.601 (.091) &      0.651 (.026) & 0.342 (.019) &  0.350 (.034) & 0.438 (.034) & 0.386 (.006) &    0.281 (.038) \\
        isomers\_c7 & 0.239 (.077) &      0.366 (.043) & 0.854 (.015) &  0.325 (.028) & 0.168 (.034) & 0.161 (.017) &    0.673 (.030) \\
  isomers\_c9 & 0.049 (.015) &      0.110 (.031) & 0.657 (.020) &  0.200 (.030) & 0.106 (.021) & 0.084 (.009) &    0.345 (.145) \\
                    JNK3 & 0.469 (.138) &      0.440 (.022) & 0.219 (.021) &  0.208 (.022) & 0.238 (.024) & 0.241 (.026) &    0.154 (.018) \\
                 median1 & 0.171 (.009) &      0.202 (.004) & 0.180 (.009) &  0.201 (.003) & 0.205 (.005) & 0.202 (.006) &    0.178 (.009) \\
                 median2 & 0.182 (.006) &      0.180 (.000) & 0.121 (.005) &  0.185 (.001) & 0.200 (.004) & 0.195 (.001) &    0.179 (.004) \\
    mestranol& 0.370 (.014) &      0.322 (.007) & 0.371 (.016) &  0.386 (.009) & 0.409 (.019) & 0.399 (.005) &    0.361 (.016) \\
         osimertinib& 0.750 (.012) &      0.784 (.001) & 0.672 (.027) &  0.765 (.002) & 0.764 (.001) & 0.771 (.002) &    0.749 (.007) \\
         perindopril& 0.432 (.013) &      0.430 (.010) & 0.172 (.088) &  0.429 (.003) & 0.445 (.004) & 0.442 (.004) &    0.421 (.008) \\
                     QED & 0.926 (.003) &      0.921 (.004) & 0.860 (.014) &  0.936 (.001) & 0.938 (.000) & 0.938 (.000) &    0.931 (.002) \\
          ranolazine& 0.689 (.015) &      0.652 (.002) & 0.555 (.015) &  0.452 (.025) & 0.411 (.010) & 0.457 (.012) &    0.347 (.012) \\
            scaffold\_hop & 0.489 (.010) &      0.463 (.002) & 0.413 (.009) &  0.455 (.004) & 0.471 (.002) & 0.470 (.003) &    0.456 (.003) \\
         sitagliptin& 0.009 (.005) &      0.008 (.003) & 0.281 (.022) &  0.084 (.015) & 0.022 (.003) & 0.023 (.004) &    0.088 (.013) \\
 thiothixene& 0.314 (.015) &      0.285 (.012) & 0.223 (.029) &  0.297 (.004) & 0.317 (.003) & 0.317 (.007) &    0.288 (.006) \\
troglitazone& 0.259 (.016) &      0.188 (.001) & 0.152 (.013) &  0.243 (.004) & 0.249 (.003) & 0.257 (.003) &    0.240 (.002) \\
        valsartan\_smarts & 0.000 (.000) &      0.000 (.000) & 0.000 (.000) &  0.002 (.003) & 0.000 (.000) & 0.002 (.004) &    0.006 (.012) \\
            zaleplon& 0.049 (.027) &      0.035 (.030) & 0.244 (.015) &  0.206 (.015) & 0.072 (.014) & 0.039 (.012) &    0.091 (.013) \\
                 Average & 0.425 (.278) &      0.397 (.247) & 0.389 (.227) &  0.386 (.217) & 0.375 (.234) & 0.373 (.238) &    0.372 (.226) \\
      Rank by avg. score &            16 &                 17 &            18 &             19 &            20 &            21 &               22 \\
\bottomrule
\end{tabular}
\label{tab_bsl_comp_gao_p3}
\end{table}

\begin{table}[h]
\tiny
\centering
\caption{Comparison of \textsc{GP-MoLFormer-Sim+GA} to main results in \cite{gao2022sample} - Page 4 of 4. Mean ($\pm$ standard deviation) AUC top-10 over 5 runs for each.}
\begin{tabular}{lcccc}
\toprule
                    Task & GFlowNet-AL Fragments & JT-VAE BO Fragments & Graph MCTS Atoms &  MolDQN Atoms \\
\midrule
    albuterol&         0.390 (.008) &       0.485 (.029) &    0.580 (.023) & 0.320 (.015) \\
          amlodipine&         0.428 (.002) &       0.519 (.009) &    0.447 (.008) & 0.311 (.008) \\
   celecoxib&         0.257 (.003) &       0.299 (.009) &    0.264 (.013) & 0.099 (.005) \\
                deco\_hop &         0.583 (.001) &       0.585 (.002) &    0.554 (.002) & 0.546 (.001) \\
                    DRD2 &         0.468 (.046) &       0.506 (.136) &    0.300 (.050) & 0.025 (.001) \\
        fexofenadine&         0.688 (.002) &       0.667 (.010) &    0.574 (.009) & 0.478 (.012) \\
                    GSK3 &         0.588 (.015) &       0.350 (.051) &    0.281 (.022) & 0.241 (.008) \\
        isomers\_c7 &         0.241 (.055) &       0.103 (.016) &    0.530 (.035) & 0.431 (.035) \\
  isomers\_c9 &         0.064 (.012) &       0.090 (.035) &    0.454 (.067) & 0.342 (.026) \\
                    JNK3 &         0.362 (.021) &       0.222 (.009) &    0.110 (.019) & 0.111 (.008) \\
                 median1 &         0.190 (.002) &       0.179 (.003) &    0.195 (.005) & 0.122 (.007) \\
                 median2 &         0.173 (.001) &       0.180 (.003) &    0.132 (.002) & 0.088 (.003) \\
    mestranol&         0.295 (.004) &       0.356 (.013) &    0.281 (.008) & 0.188 (.007) \\
         osimertinib&         0.787 (.003) &       0.775 (.004) &    0.700 (.004) & 0.674 (.006) \\
         perindopril&         0.421 (.002) &       0.430 (.009) &    0.277 (.013) & 0.213 (.043) \\
                     QED &         0.902 (.005) &       0.934 (.002) &    0.892 (.006) & 0.731 (.018) \\
          ranolazine&         0.632 (.007) &       0.508 (.055) &    0.239 (.027) & 0.051 (.020) \\
            scaffold\_hop &         0.460 (.002) &       0.470 (.005) &    0.412 (.003) & 0.405 (.004) \\
         sitagliptin&         0.006 (.001) &       0.046 (.027) &    0.056 (.012) & 0.003 (.002) \\
 thiothixene&         0.266 (.005) &       0.282 (.008) &    0.231 (.004) & 0.099 (.007) \\
troglitazone&         0.186 (.003) &       0.237 (.005) &    0.224 (.009) & 0.122 (.004) \\
        valsartan\_smarts &         0.000 (.000) &       0.000 (.000) &    0.000 (.000) & 0.000 (.000) \\
            zaleplon&         0.010 (.001) &       0.125 (.038) &    0.058 (.019) & 0.010 (.005) \\
                 Average &         0.365 (.246) &       0.363 (.236) &    0.339 (.219) & 0.244 (.211) \\
      Rank by avg. score &                    23 &                  24 &               25 &            26 \\
\bottomrule
\end{tabular}
\label{tab_bsl_comp_gao_p4}
\end{table}

\begin{table}[h]
\centering
\tiny
\caption{Comparison of \textsc{GP-MoLFormer-Sim+GA} to main results in \cite{kim2024geneticguidedgflownetssampleefficient} - Page 1 of 2. Mean ($\pm$ standard deviation) AUC top-10 over 5 runs for each.}
\begin{tabular}{lcccccc}
\toprule
                    Task &  Our Rank &       Our AUC &   Genetic GFN &        Mol GA & SMILES REINVENT &          GEGL \\
\midrule
    albuterol&         6 & 0.824 (.071) & 0.949 (.010) & 0.928 (.015) &   0.881 (.016) & 0.842 (.019) \\
          amlodipine&         3 & 0.680 (.064) & 0.761 (.019) & 0.740 (.055) &   0.644 (.019) & 0.626 (.018) \\
   celecoxib&         4 & 0.716 (.067) & 0.802 (.029) & 0.629 (.062) &   0.717 (.027) & 0.699 (.041) \\
                deco\_hop &         2 & 0.710 (.058) & 0.733 (.109) & 0.656 (.013) &   0.662 (.044) & 0.656 (.039) \\
                    DRD2 &         3 & 0.956 (.010) & 0.974 (.006) & 0.950 (.004) &   0.957 (.007) & 0.898 (.015) \\
        fexofenadine&         3 & 0.798 (.028) & 0.856 (.039) & 0.835 (.012) &   0.781 (.013) & 0.769 (.009) \\
                    GSK3 &         1 & 0.896 (.035) & 0.881 (.042) & 0.894 (.025) &   0.885 (.031) & 0.816 (.027) \\
        isomers\_c7 &         3 & 0.932 (.011) & 0.969 (.003) & 0.926 (.014) &   0.942 (.012) & 0.930 (.011) \\
  isomers\_c9 &         3 & 0.864 (.016) & 0.897 (.007) & 0.894 (.005) &   0.838 (.030) & 0.808 (.007) \\
                    JNK3 &         2 & 0.806 (.087) & 0.764 (.069) & 0.835 (.040) &   0.782 (.029) & 0.580 (.086) \\
                 median1 &         3 & 0.340 (.034) & 0.379 (.010) & 0.329 (.006) &   0.363 (.011) & 0.338 (.016) \\
                 median2 &         6 & 0.255 (.031) & 0.294 (.007) & 0.284 (.035) &   0.281 (.002) & 0.274 (.007) \\
    mestranol&         3 & 0.658 (.118) & 0.708 (.057) & 0.762 (.048) &   0.634 (.042) & 0.599 (.035) \\
         osimertinib&         5 & 0.819 (.004) & 0.860 (.008) & 0.853 (.005) &   0.834 (.010) & 0.832 (.005) \\
         perindopril&         3 & 0.584 (.042) & 0.595 (.014) & 0.610 (.038) &   0.535 (.015) & 0.537 (.015) \\
                     QED &         5 & 0.940 (.001) & 0.942 (.000) & 0.941 (.001) &   0.941 (.000) & 0.941 (.001) \\
          ranolazine&         3 & 0.812 (.024) & 0.819 (.018) & 0.830 (.010) &   0.770 (.005) & 0.730 (.011) \\
            scaffold\_hop &         5 & 0.531 (.016) & 0.615 (.100) & 0.568 (.017) &   0.551 (.024) & 0.531 (.010) \\
         sitagliptin&         3 & 0.501 (.081) & 0.634 (.039) & 0.677 (.055) &   0.470 (.041) & 0.402 (.024) \\
 thiothixene&         6 & 0.504 (.033) & 0.583 (.034) & 0.544 (.067) &   0.544 (.026) & 0.515 (.028) \\
troglitazone&         4 & 0.437 (.067) & 0.511 (.054) & 0.487 (.024) &   0.458 (.018) & 0.420 (.031) \\
        valsartan\_smarts &         2 & 0.158 (.317) & 0.135 (.271) & 0.000 (.000) &   0.182 (.363) & 0.119 (.238) \\
            zaleplon&         4 & 0.504 (.022) & 0.552 (.033) & 0.514 (.033) &   0.533 (.009) & 0.492 (.021) \\
                 Average & 3.6 (1.3) & 0.662 (.221) & 0.705 (.219) & 0.682 (.239) &   0.660 (.213) & 0.624 (.215) \\
      Rank by avg. score &        -- &             3 &             1 &             2 &               4 &             5 \\
\bottomrule
\end{tabular}
\label{tab:bsl_comp_gflow_p1}
\end{table}

\begin{table}[h]
\centering
\tiny
\caption{Comparison of \textsc{GP-MoLFormer-Sim+GA} to main results in \cite{kim2024geneticguidedgflownetssampleefficient} - Page 2 of 2. Mean ($\pm$ standard deviation) AUC top-10 over 5 runs for each.}
\begin{tabular}{lccc}
\toprule
                    Task &         GP BO &  Fragment GFN & Fragment GFN-AL \\
\midrule
    albuterol& 0.902 (.011) & 0.382 (.010) &   0.459 (.028) \\
          amlodipine& 0.579 (.035) & 0.428 (.002) &   0.437 (.007) \\
   celecoxib& 0.746 (.025) & 0.263 (.009) &   0.326 (.008) \\
                deco\_hop & 0.615 (.009) & 0.582 (.001) &   0.587 (.002) \\
                    DRD2 & 0.941 (.017) & 0.480 (.075) &   0.601 (.055) \\
        fexofenadine& 0.726 (.004) & 0.689 (.003) &   0.700 (.005) \\
                    GSK3 & 0.861 (.027) & 0.589 (.009) &   0.666 (.006) \\
        isomers\_c7 & 0.883 (.040) & 0.791 (.024) &   0.468 (.211) \\
  isomers\_c9 & 0.805 (.007) & 0.576 (.021) &   0.199 (.199) \\
                    JNK3 & 0.611 (.080) & 0.359 (.009) &   0.442 (.017) \\
                 median1 & 0.298 (.016) & 0.192 (.003) &   0.207 (.003) \\
                 median2 & 0.296 (.011) & 0.174 (.002) &   0.181 (.002) \\
    mestranol& 0.631 (.093) & 0.291 (.005) &   0.332 (.012) \\
         osimertinib& 0.788 (.005) & 0.787 (.002) &   0.785 (.003) \\
         perindopril& 0.494 (.006) & 0.423 (.006) &   0.434 (.006) \\
                     QED & 0.937 (.002) & 0.904 (.002) &   0.917 (.002) \\
          ranolazine& 0.741 (.010) & 0.626 (.005) &   0.660 (.004) \\
            scaffold\_hop & 0.535 (.007) & 0.461 (.002) &   0.464 (.003) \\
         sitagliptin& 0.461 (.057) & 0.180 (.012) &   0.217 (.022) \\
 thiothixene& 0.544 (.038) & 0.261 (.004) &   0.292 (.009) \\
troglitazone& 0.404 (.025) & 0.183 (.001) &   0.190 (.002) \\
        valsartan\_smarts & 0.000 (.000) & 0.000 (.000) &   0.000 (.000) \\
            zaleplon& 0.466 (.025) & 0.308 (.027) &   0.353 (.024) \\
                 Average & 0.620 (.232) & 0.432 (.227) &   0.431 (.219) \\
      Rank by avg. score &             6 &             7 &               8 \\
\bottomrule
\end{tabular}
\label{tab:bsl_comp_gflow_p2}
\end{table}

\begin{table}[h]
\tiny
\centering
\caption{Comparison of \textsc{GP-MoLFormer-Sim+GA} to main results in \cite{wang2024efficient}. Mean ($\pm$ standard deviation) AUC top-10 over 5 runs for each.}
\begin{tabular}{lccccccccc}
\toprule
                    Task &  Our  &       Our  &      REINVENT & Augmented  &  Graph  &         GP BO & MOLLEO & MOLLEO  & MOLLEO  \\
                         &   Rank &        AUC &              &  Memory &         GA &         GP BO & (MolSTM) & (BioT5) &(GPT-4) \\
\midrule
    albuterol&         8 & 0.824 (.071) & 0.896 (.008) &    0.918 (.026) & 0.874 (.020) & 0.902 (.019) &   0.929 (.005) &  0.968 (.003) &  0.985 (.024) \\
          amlodipine&         4 & 0.680 (.064) & 0.642 (.044) &    0.686 (.046) & 0.625 (.040) & 0.552 (.025) &   0.674 (.018) &  0.776 (.038) &  0.773 (.037) \\
   celecoxib&         5 & 0.716 (.067) & 0.716 (.084) &    0.784 (.011) & 0.582 (.057) & 0.728 (.048) &   0.594 (.105) &  0.508 (.017) &  0.864 (.034) \\
                deco\_hop &         3 & 0.710 (.058) & 0.666 (.044) &    0.688 (.060) & 0.619 (.004) & 0.629 (.018) &   0.613 (.016) &  0.827 (.093) &  0.942 (.013) \\
                    DRD2 &         6 & 0.956 (.010) & 0.945 (.007) &    0.962 (.005) & 0.964 (.012) & 0.923 (.017) &   0.975 (.003) &  0.981 (.002) &  0.968 (.012) \\
        fexofenadine&         2 & 0.798 (.028) & 0.769 (.009) &    0.686 (.010) & 0.779 (.025) & 0.745 (.009) &   0.789 (.016) &  0.773 (.017) &  0.847 (.018) \\
                    GSK3 &         2 & 0.896 (.035) & 0.865 (.043) &    0.889 (.027) & 0.788 (.070) & 0.851 (.041) &   0.898 (.041) &  0.889 (.015) &  0.863 (.047) \\
        isomers\_c7 &         5 & 0.932 (.011) & 0.842 (.029) &    0.954 (.033) & 0.949 (.036) & 0.662 (.071) &   0.948 (.036) &  0.928 (.038) &  0.984 (.008) \\
  isomers\_c9 &         4 & 0.864 (.016) & 0.642 (.054) &    0.830 (.016) & 0.719 (.047) & 0.469 (.180) &   0.871 (.039) &  0.873 (.019) &  0.874 (.053) \\
                    JNK3 &         1 & 0.806 (.087) & 0.783 (.023) &    0.773 (.073) & 0.553 (.136) & 0.564 (.155) &   0.643 (.226) &  0.728 (.079) &  0.790 (.027) \\
                 median1 &         3 & 0.340 (.034) & 0.372 (.015) &    0.335 (.012) & 0.287 (.008) & 0.325 (.012) &   0.298 (.019) &  0.338 (.033) &  0.352 (.024) \\
                 median2 &         6 & 0.255 (.031) & 0.294 (.006) &    0.290 (.006) & 0.229 (.017) & 0.308 (.034) &   0.251 (.031) &  0.259 (.019) &  0.275 (.045) \\
    mestranol&         4 & 0.658 (.118) & 0.618 (.048) &    0.764 (.035) & 0.579 (.022) & 0.627 (.089) &   0.596 (.018) &  0.717 (.104) &  0.972 (.009) \\
         osimertinib&         5 & 0.819 (.004) & 0.834 (.046) &    0.856 (.013) & 0.808 (.012) & 0.762 (.029) &   0.823 (.007) &  0.817 (.016) &  0.835 (.024) \\
         perindopril&         4 & 0.584 (.042) & 0.537 (.016) &    0.598 (.008) & 0.538 (.009) & 0.493 (.011) &   0.554 (.037) &  0.738 (.016) &  0.600 (.031) \\
                     QED &         4 & 0.940 (.001) & 0.941 (.000) &    0.941 (.000) & 0.940 (.000) & 0.937 (.000) &   0.937 (.002) &  0.937 (.002) &  0.948 (.000) \\
          ranolazine&         1 & 0.812 (.024) & 0.760 (.009) &    0.802 (.003) & 0.728 (.012) & 0.735 (.013) &   0.725 (.040) &  0.749 (.012) &  0.769 (.022) \\
            scaffold\_hop &         6 & 0.531 (.016) & 0.560 (.019) &    0.565 (.008) & 0.517 (.007) & 0.548 (.019) &   0.527 (.019) &  0.559 (.102) &  0.971 (.004) \\
         sitagliptin&         4 & 0.501 (.081) & 0.021 (.003) &    0.479 (.039) & 0.433 (.075) & 0.186 (.055) &   0.548 (.065) &  0.506 (.100) &  0.584 (.067) \\
 thiothixene&         7 & 0.504 (.033) & 0.534 (.013) &    0.562 (.028) & 0.479 (.025) & 0.559 (.027) &   0.508 (.035) &  0.696 (.081) &  0.727 (.052) \\
troglitazone&         4 & 0.437 (.067) & 0.452 (.048) &    0.556 (.052) & 0.377 (.010) & 0.405 (.007) &   0.381 (.025) &  0.390 (.044) &  0.562 (.019) \\
        valsartan &         2 & 0.158 (.317) & 0.000 (.000) &    0.000 (.000) & 0.000 (.000) & 0.000 (.000) &   0.000 (.000) &  0.000 (.000) &  0.867 (.092) \\
            zaleplon&         2 & 0.504 (.022) & 0.347 (.049) &    0.438 (.082) & 0.456 (.007) & 0.272 (.026) &   0.475 (.018) &  0.465 (.026) &  0.510 (.031) \\
                 Average & 4.0 (1.8) & 0.662 (.221) & 0.610 (.260) &    0.668 (.237) & 0.601 (.239) & 0.573 (.239) &   0.633 (.245) &  0.671 (.248) &  0.777 (.200) \\
      Rank by &        -- &             4 &             6 &                3 &             7 &             8 &               5 &              2 &              1 \\
      avg. score &       &&&&&&&&\\
\bottomrule
\end{tabular}
\label{tab:bsl_comparison_wang}
\end{table}

\begin{table*}[h]
\caption{The mean ($\pm$ standard deviation) Synthetic Accessibility scores (SA) of top 100 molecules generated for each task using \textsc{GP-MoLFormer-Sim+GA}. The mean values are less than the threshold of 6 suggested in \cite{Ertl_Schuffenhauer_2009}, indicating that the molecules are easily synthesizable.}
    \centering
    \scriptsize
    \begin{tabular}{lc}
    \toprule
        Task & SA \\ 
    \midrule
        albuterol & 4.235 (0.433) \\
        amlodipine & 3.453 (0.101) \\
        celecoxib & 2.422 (0.322) \\ 
        deco\_hop & 3.831 (0.541) \\
        DRD2 & 4.004 (0.401) \\ 
        fexofenadine & 5.285 (0.320) \\
        GSK3 & 3.230 (0.425) \\ 
        isomers\_c7& 2.834 (0.795) \\ 
        isomers\_c9 & 4.301 (0.226) \\ 
        JNK3 (zincseed)+RFF768+XO+DIV & 3.114 (0.371) \\ 
        median1 (zincseed)+RFF768+XO+DIV & 3.587 (0.041) \\ 
        median2 (zincseed)+RFF768+XO+DIV & 2.882 (0.353) \\ 
        mestranol & 3.836 (0.203) \\ 
        osimertimib & 4.031 (0.155) \\ 
        perindopril & 3.791 (0.289) \\ 
        QED & 2.417 (0.314) \\
        ranolazine & 4.040 (0.292) \\ 
        scaffold\_hop & 3.621 (0.273) \\ 
        sitagliptin & 3.967 (0.460) \\ 
        thiothixene & 2.546 (0.360) \\ 
        troglitazone & 3.329 (0.228) \\ 
        valsartan & 3.066 (1.031) \\ 
        zaleplon & 3.461 (0.501) \\ 
        \bottomrule
    \end{tabular}
\label{tab:sa}
\end{table*}

\subsection{Extended Results --- Trypsin inhibitors}

\begin{figure}[!ht]
    \centering
    \includegraphics[width=1.\textwidth]{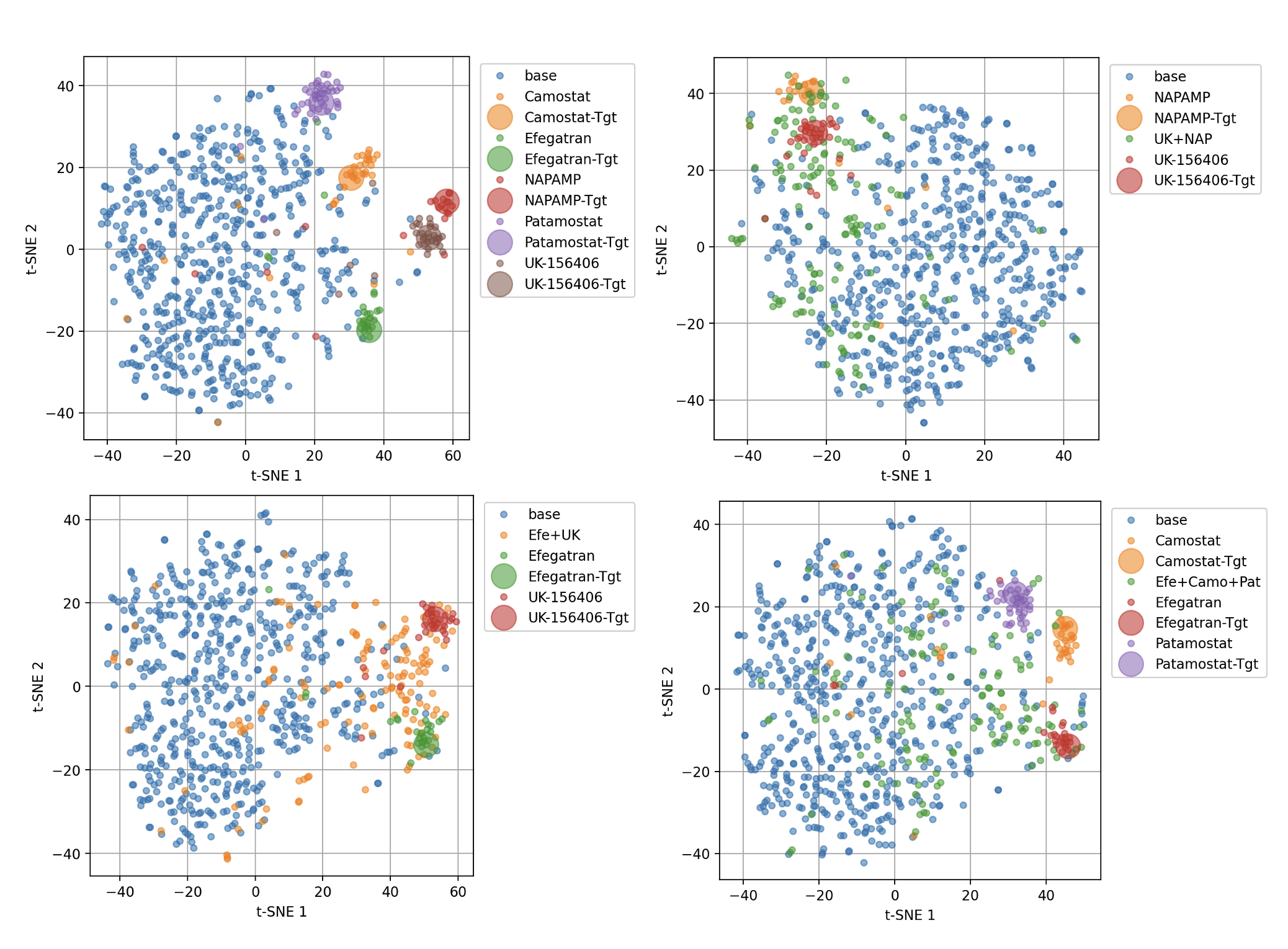}
    \caption{Examples of guided generation operating in single-guide (top left) and multi-guide mode (top right, and bottom) using selected combinations of Trypsin inhibitors.} 
    \label{fig:trypsin_multiguide_clusters}
\end{figure}

Algorithm \ref{alg:guided_generation} anticipates the possibility of guidance by multiple targets at the same time (see Line 19 of Algorithm \ref{alg:guided_generation} in Section \ref{sec:guided_generation}). In order to maintain focus on high scoring candidates in the context of the GA-based optimization, we operated the guidance mechanism in a single-target mode in main experiments. However, the guidance mechanism via \textsc{GP-MoLFormer-Sim} operates equally well in multi-guide mode. This is exemplified visually in Figure \ref{fig:trypsin_multiguide_clusters} using the t-SNE projection. The top left plot shows single-guide generations for the five target compounds for reference. The top right and bottom plots show generations guided by two of the compounds simultaneously (NAPAMP+UK-156406 and Efegatran+UK-156406, respectively). In the multi-guide cases, a clear trend in the direction of the combined targets on the t-SNE plot can be observed indicating guidance efficacy. Such mode can be of practical use in applications involving multi-objective or group-wise molecular optimization.  

\clearpage
\section{Hyperparameter Settings}
Typical values for essential hyperparameters are listed in Table \ref{tab:hyperparameters}.
\begin{table}[h]
\centering
\scriptsize
\caption{\textsc{GP-MoLFormer-Sim+GA} hyperparameters used.}
\begin{tabular}{p{0.15\linewidth}lp{0.3\linewidth}lp{0.25\linewidth}}
% \begin{tabular}{lllll}
\toprule
Parameter &  Type & Description &  Value &  Comment \\
\midrule
Multi-guide generation& Boolean & Each guide considered a separate target & True & Otherwise average of guides taken\\ 
Number of guides & Int & Best-score candidates & 3 & May be subject to pruning due to low Oracle value. Diversity guides are additional. \\
Guide pruning percentage & Float & Guides below this \% of the top guide are pruned & 75.0 & \\
Post-gen pruning $K$ & Integer & Offspring's Tanimoto to top-1 must be larger than that of the $K$-th best candidate before being sent to the Oracle & 10 & This is an exploitative step. Exploration arrangements are done separately \\
Exploration candidates & Int & Add up to this number of exploration candidates per generation & 40 & \\
Exploration method & String & Method to select exploration candidates: ``Random'' or ``Crossover'' & Varies & In ``Crossover'' mode, best guides serve as parents to create novel candidates\\
Diversity guides & Int & number of diversity guides to add in each guided run & 1 & This is in addition to parameter ``Number of guides'' above \\
Guidance strength $\alpha$ and temperature $\tau$ & Float & See Algorithm \ref{alg:guided_generation} & 0.4, 0.25 & \\
Exploitation trigger & Float & Change schedule if the top-1 Oracle value exceeds this value & 0.95 & This usually implies the target has been hit and the remaining 9 of 10 candidates should now get as close as possible to the top-1. Strategy is switched from exploration to exploitation \\
Exploitation $\alpha, \tau$ & Float & Values when exploitation mode is active & 0.4, 0.15 & Guiding at low temperature produces candidate very close to the guide\\
Stop after no change & Int & With no progress after this many generations, quit & 500 & \\
Maximum generation size & Int & At any generation cap number of candidates sent to the Oracle & 120 & \\
Generation size per guide & Int & & 20 & \\
RFF Dimension & Int & Number of RFF features & 768 & \\
RFF (Entropy) Temperature & Float & Corresponds to variance in the Gaussian kernel aproximation & 0.008 & \\
\textsc{GP-MoLFormer} embedding dimensionality & Float & & 768 & See \cite{ross2025gpmolformerfoundationmodelmolecular}\\
\bottomrule
\end{tabular}
\label{tab:hyperparameters}
\end{table}

\end{document}